\documentclass[lettersize,journal]{IEEEtran}
\usepackage{amsmath,amsfonts}
\usepackage{array}
\usepackage[caption=false,font=normalsize,labelfont=sf,textfont=sf]{subfig}
\usepackage{textcomp}
\usepackage{stfloats}
\usepackage{url}
\usepackage{verbatim}
\usepackage{graphicx}
\usepackage{multirow}
\usepackage{amsthm}

\usepackage{bm}
\usepackage{color}
\usepackage{nomencl}
\usepackage[linesnumbered,ruled,vlined]{algorithm2e}
\usepackage[hidelinks]{hyperref}

\usepackage{amsthm}
\usepackage{amssymb}
\usepackage[noend]{algpseudocode}

\usepackage{amsmath}
\usepackage{float}
\usepackage{tabularx}
\usepackage{booktabs}
\usepackage{tikz}

\makenomenclature
\def\BibTeX{{\rm B\kern-.05em{\sc i\kern-.025em b}\kern-.08em
    T\kern-.1667em\lower.7ex\hbox{E}\kern-.125emX}}
\usepackage{balance}
\begin{document}
\title{Genetic Informed Trees (GIT*): Path 
Planning via Reinforced Genetic Programming Heuristics
}
\author{Liding Zhang$^{1}$, Kuanqi Cai$^{1}$, Zhenshan Bing$^{1}$, Chaoqun Wang$^{2}$, Alois Knoll$^{1}$, ~\IEEEmembership{Fellow,~IEEE} 

\thanks{$^{1}$L. Zhang, K. Cai, Z. Bing, and A. Knoll are with the Department of Informatics, Technical University of Munich, Germany. {\tt\small liding.zhang@tum.de}}

\thanks{$^{2}$C. Wang is with the School of Control Science and Engineering, Shandong University, Shandong, China. \\
\textit{(Corresponding authors: Zhenshan Bing; Kuanqi Cai.)}}}
\markboth{IEEE TRANSACTIONS ON \LaTeX}%
{How to Use the IEEEtran \LaTeX \ Templates}

\maketitle

\begin{abstract}
Optimal path planning involves finding a feasible state sequence between a start and a goal that optimizes an objective. This process relies on heuristic functions to guide the search direction. While a robust function can improve search efficiency and solution quality, current methods often overlook available environmental data and simplify the function structure due to the complexity of information relationships. This study introduces Genetic Informed Trees (GIT*), which improves upon Effort Informed Trees (EIT*) by integrating a wider array of environmental data, such as repulsive forces from obstacles and the dynamic importance of vertices, to refine heuristic functions for better guidance.
Furthermore, we integrated reinforced genetic programming (RGP), which combines genetic programming with reward system feedback to mutate genotype-generative heuristic functions for GIT*. RGP leverages a multitude of data types, thereby improving computational efficiency and solution quality within a set timeframe. Comparative analyses demonstrate that GIT* surpasses existing single-query, sampling-based planners in problems ranging from $\mathbb{R}^4$ to $\mathbb{R}^{16}$ and was tested on a real-world mobile manipulation task.
A video showcasing our experimental results is available at \href{https://youtu.be/URjXbc_BiYg}{\textcolor{blue}{https://youtu.be/URjXbc\_BiYg}}.
\end{abstract}


\section{Introduction}
\IEEEPARstart{P}{ath} {planning is a fundamental challenge in robotic automation, involving the determination of a sequence of valid states that guide a robot from a starting point to a desired goal while avoiding obstacles~\cite{zhang2024review}. Many algorithms have been proposed to address this problem, such as the A* algorithm~\cite{hart1968formal}, Artificial Potential Field (APF) algorithm~\cite{Khatib1986}, and sampling-based algorithms~\cite{karaman2011sampling}. The A* algorithm's performance declines with higher dimensionality, while the APF algorithm often converges to local minima. Sampling-based algorithms have gained popularity due to their efficient exploration of the state space~\cite{Caisampling}.} 
However, they often require significant time to find the optimal solution. In multi-dimensional environments, such as autonomous vehicles and robot manipulators, it is essential to compute an efficient path to conserve power~\cite{zhang2025apt}.

\begin{figure}[t!] 
    \centering 
    \includegraphics[width=0.48\textwidth]{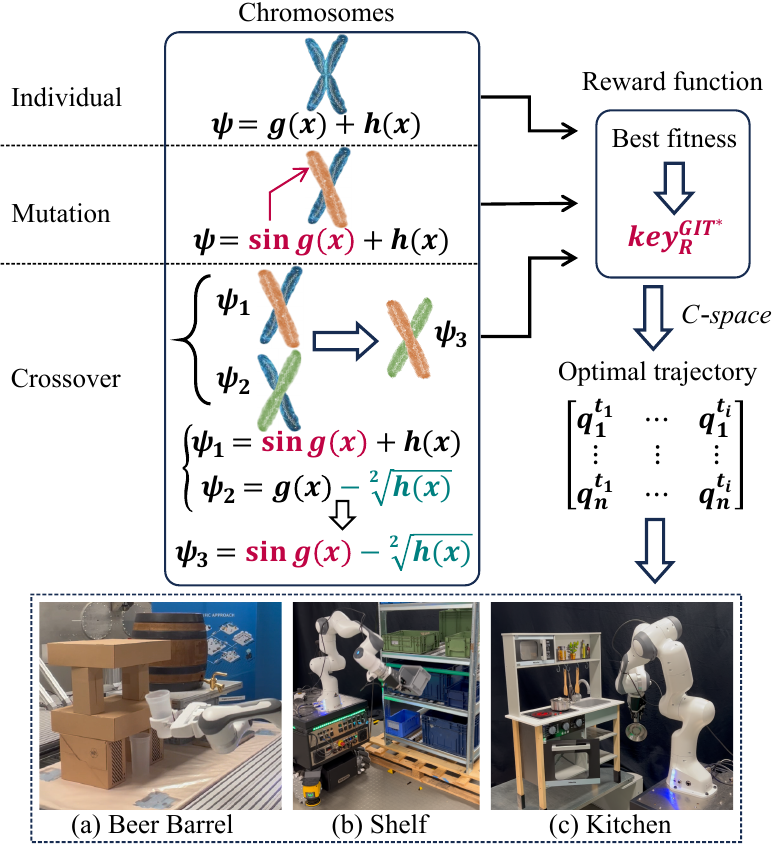} 
    \vspace{-0.5em} 
    \caption{{GIT* utilizes a population of G-heuristics (i.e., $\psi_1, \psi_2$...) with chromosome behaviors. The example of G-heuristics is illustrated above, which are trained using a reward function over multiple generations through RGP. The best G-heuristic is employed for pathfinding guidance. GIT*'s performance is evaluated in beer barrel, shelf, and kitchen model scenarios.}
}  
    \label{fig: first} 
    \vspace{-1.7em} 
\end{figure}
{The motivation for this paper is to improve the convergence rate and find successful solutions faster with lower initial solution costs based on the genetic-based generation of heuristics.
Sampling-based algorithms like Rapidly-exploring Random Trees (RRT)~\cite{lavalle2001randomized},  Probabilistic Roadmaps (PRM)~\cite{kavraki1996probabilistic}, and variant algorithms of RRTs~\cite{Zhang2024Elliptical} have been widely used for recent path planning work and have demonstrated effectiveness in practical applications. However, these algorithms' performance fluctuates greatly in different environments.} On the other hand, in optimization algorithm research, a combined method known as Reinforced Genetic Programming (RGP)~\cite{Downing2001} is proposed. We introduced \textit{genotype-generative heuristic} (G-heuristic) functions based on RGP for optimal edge evaluation, incorporating a fitness reward function to facilitate autonomous learning and adjustment of exploration strategies based on environmental feedback. This approach integrates the Genetic Algorithm (GA)~\cite{Sastry2005} to assess bio-inspired chromosome behavior (e.g., \textit{crossover, mutation, reproduction}) for integration with sampling-based planners. However, the G-heuristic cannot be directly applied to robot path planning because it does not consider environmental constraints (e.g., \textit{obstacle avoidance}) or robustness across various scenarios. Therefore, the G-heuristic must be trained across different benchmarking datasets using reward feedback for robustness.

\begin{figure}[t!]
    \centering
    \begin{tikzpicture}
    \node[inner sep=0pt] (russell) at (-3,0)
    {\includegraphics[width=0.24\textwidth]{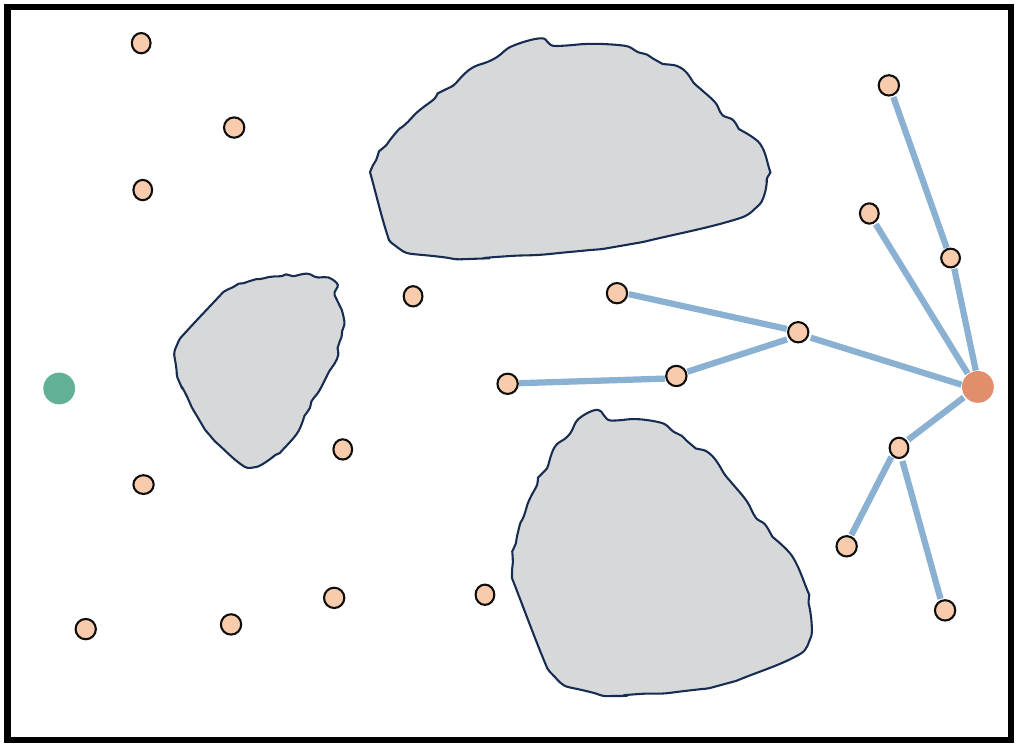}};
    \node[inner sep=0pt] (russell) at (1.4,0)
    {\includegraphics[width=0.24\textwidth]{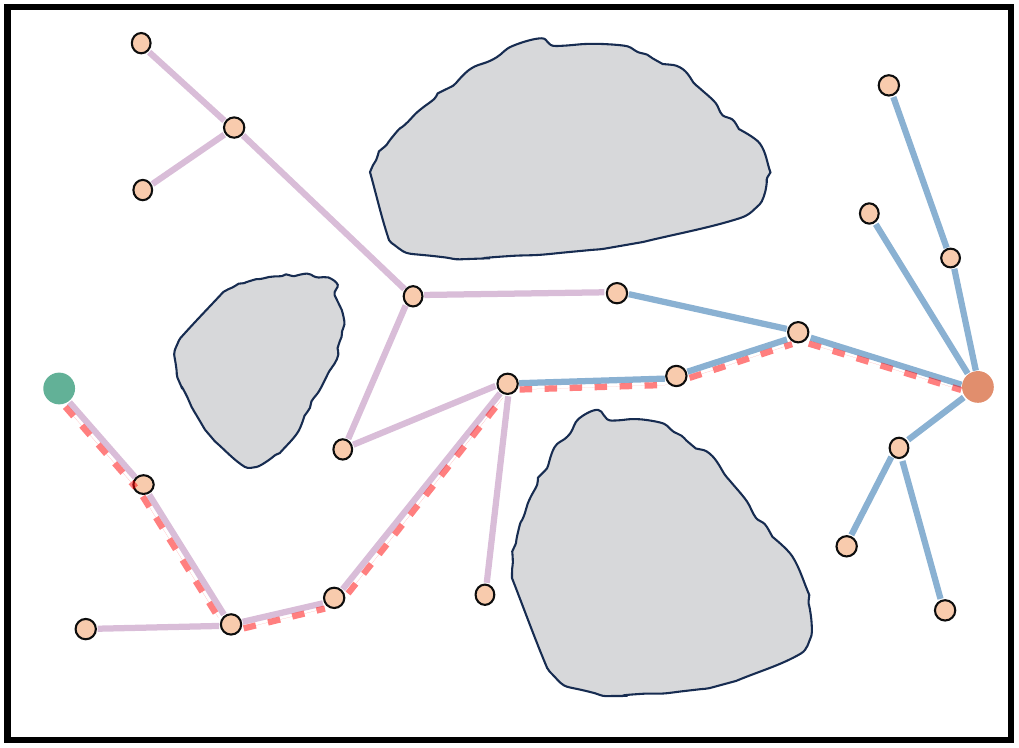}};
    \node[inner sep=0pt] (russell) at (-3,-3.7)
    {\includegraphics[width=0.24\textwidth]{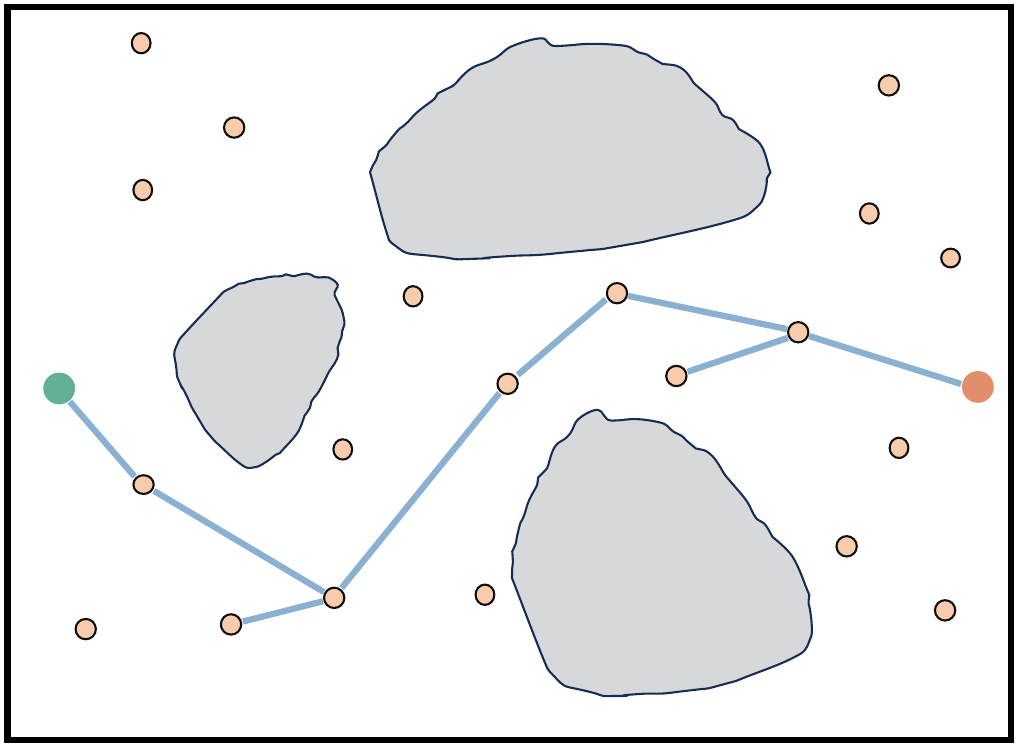}};
    \node[inner sep=0pt] (russell) at (1.4,-3.7)
    {\includegraphics[width=0.24\textwidth]{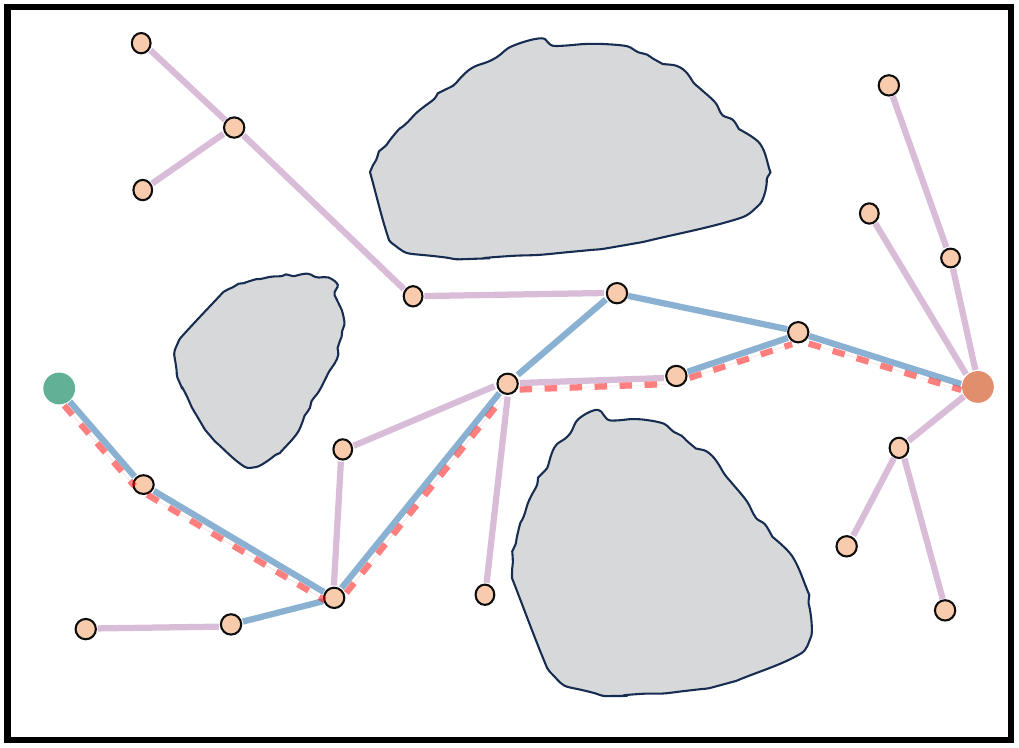}};

    \node at (-4.85,0.2) {\color{teal}\scriptsize start};
    \node at (-0.45,-3.5) {\color{teal}\scriptsize start};
    \node at (-0.45,0.2) {\color{teal}\scriptsize start};
    \node at (-4.85,-3.5) {\color{teal}\scriptsize start};

    \node at (2.95,-0.12) {\color{purple}\scriptsize goal};
    \node at (-1.45,-3.82) {\color{purple}\scriptsize goal};
    \node at (-1.45,-0.12) {\color{purple}\scriptsize goal};
    \node at (2.95,-3.82) {\color{purple}\scriptsize goal};
    \node at (1.6,1.0) {\small${X}_\text{obs}$};
    \node at (1.6,-2.7) {\small${X}_\text{obs}$};
    \node at (-2.8,-2.7) {\small${X}_\text{obs}$};
    \node at (-2.8,1.0) {\small${X}_\text{obs}$};

    \node at (-3,-1.8) {\small(a) EIT*  $(t_n)$};
    \node at (1.4,-1.8) {\small(b) EIT* $(t_{n+1})$};
    \node at (-3,-5.6) {\small(c) GIT* $(t_n)$};
    \node at (1.4,-5.6) {\small(d) GIT* $(t_{n+1})$};
    
    \end{tikzpicture}
    \vspace{-1.0em} 
    \caption{{Four snapshots show EIT* and GIT* exploration strategies in reverse search. GIT* employs G-heuristics, while EIT* maintains a linear combination heuristic. Yellow points indicate valid samples in obstacle-free areas. The blue line represents the reverse tree at time $t_n$, the pink line represents it at time $t_{n+1}$, and the dashed red lines are optimized reverse edges.
}}

    \label{fig: greedy}
    \vspace{-1.0em} 
\end{figure}
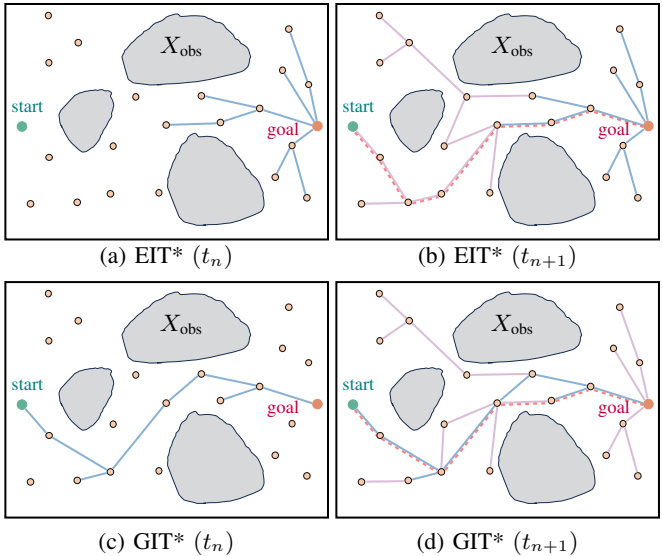

Inspired by RGP technology, this paper presents the Genetic Informed Trees (GIT*) algorithm, which generates a heuristic function using problem-specific information via RGP. This heuristic enhances efficiency by minimizing expanded vertices. GIT* uses invalid samples within obstacles and start/goal points to create an APF, incorporating obstacle shapes and locations, and tracks sample visit frequency to account for the dynamic importance of states. The G-heuristics represent a symbolic regression problem tackled by RGP. It involves evolving nonlinear expressions to refine the heuristic. As shown in Fig.~\ref{fig: greedy}, G-heuristic enables GIT* to find the initial solution quickly and then expand. GIT* incorporates additional graph search techniques, such as truncation and inflation, to balance exploitation and exploration, dynamically modified using RGP. {GIT* has shown improvements over state-of-the-art (SOTA) methods in time to find the initial solution, initial solution quality, and final solution quality in both generalized simulation benchmarks and real-world experiments.}

The contributions of this paper are summarized as follows:
\begin{enumerate}
    \item An efficient optimal genotype-generative heuristic function based on reinforced genetic programming, trained with a dataset from the random problem domain.
    \item A novel sampling-based path planning algorithm, GIT*, integrates the trained genotype-generative heuristic function to rapidly obtain high-quality solutions.
    \item {Demonstrating the effectiveness of GIT* across various dimensional environments and optimization objectives.}

\end{enumerate}

\section{Related work}
{Heuristic functions optimize path planning by estimating goal-state costs, which is crucial across multiple dimensions.} Informed planners with heuristics outperform their uninformed counterparts~\cite{gammell2018informed}. Effective heuristics should be accurate and efficient, yet balancing these traits can be challenging~\cite{strub2022adaptively}.

{RRT-Connect~\cite{kuffner2000rrt} extends the RRT framework by growing two trees: one from the start state and the other from the goal, using heuristic-guided planning to accelerate path convergence. However, RRT-Connect lacks asymptotic optimality and does not improve solution quality with more computation time~\cite{zhang25g3t}. Cost heuristics in tree growth are demonstrated by Heuristically-guided RRT (hRRT)~\cite{urmson2003approaches} and Generalized Bidirectional RRT (GBRRT)~\cite{nayak2020bidirectional}. hRRT uses a priori heuristics for exploration within RRT's Voronoi regions, while GBRRT, a bidirectional RRT variant, employs reverse tree-computed heuristics to guide the forward tree. However, these algorithms do not provide bounds on solution quality~\cite{zhang2025TASE}.}

{To overcome this limitation, some planners combine graph-based and sampling-based approaches. Batch Informed Trees (BIT*)~\cite{gammell2020batch} uses A* on a \textit{random geometric graph} (RGG) formed by random samples, improving approximation as samples increase. While BIT* efficiently refines its search, it still has limitations in using problem-specific information.
Advanced BIT* (ABIT*)~\cite{strub2020advanced} enhances BIT* by introducing inflated and truncated factors, balancing the exploration and exploitation.
Adaptively Informed Trees (AIT*)~\cite{strub2022adaptively} and Effort
Informed Trees (EIT*)~\cite{strub2022adaptively} further enhance efficiency with bidirectional search strategies. EIT* uses adaptive sparse collision checks, reducing expensive collision detections. The forward and reverse trees inform each other, sharing complementary information to optimize the search~\cite{Zhang2024Elliptical}. However, EIT* did not leverage the information from the invalid sample/nearest neighbor in the planning domain. GIT* integrates APF and dynamic importance for further guidance.}
%
%
\subsection{Genetic-based Path Planning Method}

Genetic sampling-based algorithms utilize genetic operations like crossover and mutation to generate candidate solutions in the problem space.
Hybridizing-RRT~\cite{hybridizingRRT2011} uses a hybrid path generation scheme that combines RRT with an island parallel genetic algorithms (GA) to efficiently find $G^3\text{-continuous }\eta^3\text{-spline}$ paths that optimize path length and curvature. This approach leverages RRT injections to maintain genetic diversity and prevent premature convergence in complex map scenarios.
Genetic-RRT~\cite{Wang2023} uses GA to optimize paths planned by RRTs. This approach retains multiple optimal solutions, increasing the likelihood of finding an asymptotically optimal path with more iterations.
However, genetic-based algorithms typically use GA to optimize path length objectives, often overlooking edge evaluation during exploration, resulting in a challenging achievement of rapid convergence of the initial solution. Our method enhances search efficiency by employing RGP to create G-heuristics.
\subsection{Applications of Symbolic Regression}
Symbolic regression, a popular application of genetic algorithms (GA)~\cite{rudolph1994convergence}, discovers mathematical expressions that accurately represent datasets without presupposing a specific mathematical form. Unlike traditional regression models that require predefined functional relationships, symbolic regression explores all possible expressions, uncovering complex data relationships, nonlinear interactions, and dynamic patterns~\cite{rudolph1997convergence}. 
Inspired by symbolic regression, our work incorporates genetic programming to generate heuristic functions within the GIT* algorithm for path planning. This integration allows GIT* to leverage a broader range of data, improving computational efficiency and solution quality.

The Open Motion Planning Library (OMPL) \cite{sucan2012open} is commonly used in benchmarking motion planning algorithms. It provides a comprehensive framework and tools for researchers to evaluate algorithms. Genetic Informed Trees (GIT*) is integrated into the OMPL framework, the Planner-Arena benchmark database \cite{moll2015benchmarking}, and Planner Developer Tools (PDT) \cite{gammell2022planner}.

\section{Problem Formulation}
{We define the optimal planning problem according to the definition provided in~\cite{karaman2011sampling} and consider the symbolic regression problem defined in~\cite{virgolin2022symbolic} as a tool to address optimal planning.}

\textit{Problem Definition 1 (Optimal Planning):} {Consider a path planning problem with the $n$-th dimensional state space $X \subseteq \mathbb{R}^n$. Let $X_{\text{obs}} \subset X$ represent states in collision with obstacles, and $X_{\text{free}} = cl(X \setminus X_{\text{obs}})$ denote the resulting permissible states, where $cl(\cdot)$ represents the \textit{closure} of a set. The initial/start state is denoted by $\mathbf{x}_{\text{start}} \in X_{\text{free}}$, and the set of desired final/goal states is $X_{\text{goal}} \subset X_{\text{free}}$. A sequence of states $\sigma: [0, 1] \mapsto X$ forms a continuous map (i.e., a collision-free, feasible path), and $\Sigma$ represents the set of all nontrivial paths.}

{The optimal solution, represented as the queue vector $\sigma^*$, corresponds to the path that minimizes a selected scalar cost function $s: \Sigma \mapsto \mathbb{R}_{\geq 0}$. This path connects the initial state $\mathbf{x}_{\text{start}}$ to any goal state $\mathbf{x}_{\text{goal}} \in X_{\text{goal}}$ through the free space:}
\begin{equation}
\begin{split}
    \sigma^* &= \arg \min_{\sigma \in \Sigma} \left\{ s(\sigma) \ \middle|\ \sigma(0) = \mathbf{x}_{\text{start}}, \sigma(1) \in X_{\text{goal}}, \right. \\
    &\qquad\qquad \left. \forall t \in [0, 1], \sigma(t) \in X_{\text{free}} \right\},
\end{split}
\end{equation}
{where $\mathbb{R}_{\geq 0}$ denotes non-negative real numbers. The cost of the optimal path is $s^*$, and $t$ is the timestep of the exploration.}

Considering a discrete set of states, $X_{\text{samples}} \subset X$, as a graph where edges are determined algorithmically by a transition function, we can describe its properties using a probabilistic model implicit dense RGGs when these states are randomly sampled, i.e., $X_{\text{samples}} = \{ \mathbf{x} \sim \mathcal{U}(X) \}$, as discussed in~\cite{penrose2003random}.

The characteristics of the anytime almost-surely sampling-based planner with the definition are provided in~\cite{gammell2018informed}.

\begin{figure*}[t!] 
    \centering 
    \includegraphics[width = 0.98\textwidth]{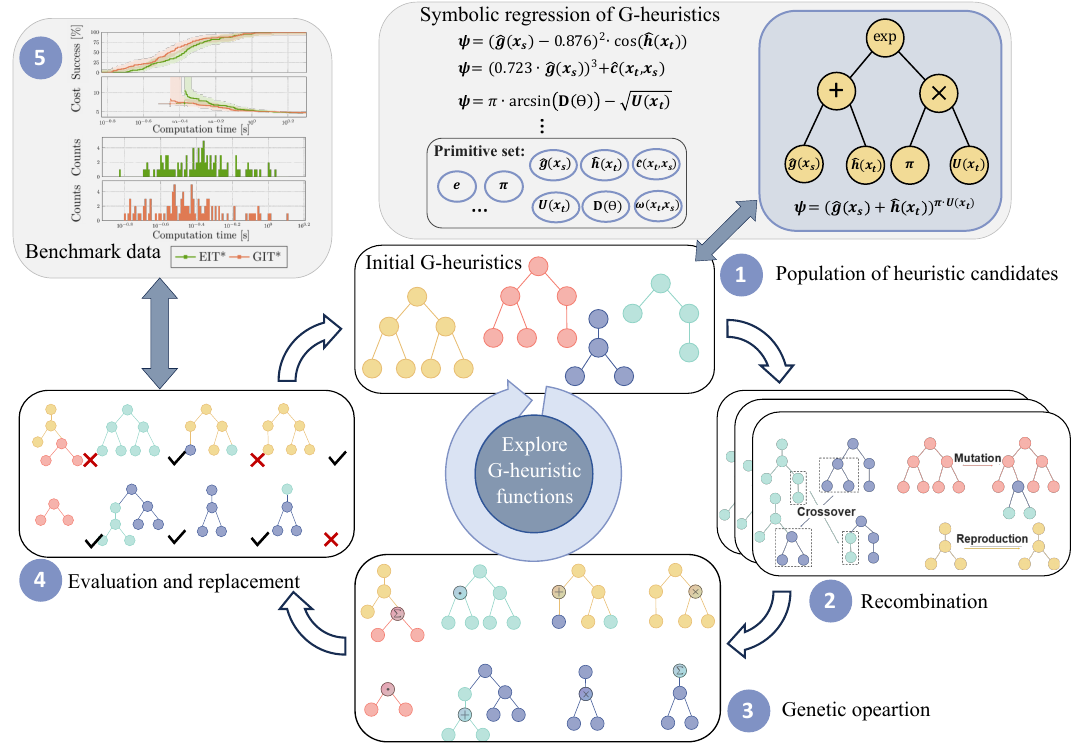}   
    \vspace{-1.0em} 
    \caption{Overview of the Reinforced Genetic Programming (RGP) process for path planning. The process initiates with the collection of path planning scenarios as benchmark data, followed by the generation of expression trees acting as G-heuristics. These trees represent individuals' heuristics in the genetic process, undergoing various genetic operations such as mutation, crossover, and reproduction to evolve a new population of heuristic candidates. The cycle concludes with the evaluation and replacement of individuals, continuously iterating to enhance algorithmic performance. } 
    \label{fig:GenerationIndividual} 
    \vspace{-1.7em} 
\end{figure*}
\textit{Problem Definition 2 (Symbolic Regression):} Symbolic regression aims to find a mathematical expression that best fits a given dataset. The process involves searching the space of mathematical expressions to identify the one that minimizes the error between the predicted output $\hat{y}$ and the actual output $y$ over a dataset $\mathcal{D}$ \cite{koza1992genetic}. This can be formulated as an optimization problem where the objective is to minimize the sum of squared errors, represented by the fitness function.

{The fitness function, $\textit{Fitness}(\cdot)$, quantifies the error between the predicted and actual outputs. It is defined as:
\begin{equation}
    \textit{Fitness}(\psi) = \sum_{(\rho_\psi, y) \in \mathcal{D}} \left( y - \hat{y}(\rho_\psi, \psi) \right)^2,
\end{equation}
where $\psi$ represents the symbolic expression, $\rho_\psi$ denotes the input fitness value, $\hat{y}(\rho_\psi, \psi)$ is the predicted output generated by the symbolic expression $\psi$, and $y$ is the actual output in the dataset $\mathcal{D}$.} The goal is to find the expression $\psi^*$ that minimizes  $\textit{Fitness}(\psi)$, thus minimizing the sum of squared errors over the dataset. This optimization problem can be expressed as:
\begin{equation}
    \psi^* = \arg \min_{\psi} \textit{Fitness}(\psi).
\end{equation}

{In this context, the fitness function measures how well a given symbolic expression $\psi$ fits the dataset $\mathcal{D}$. By minimizing the fitness function, we aim to find the symbolic expression that best fits the data, as discussed in~\cite{virgolin2022symbolic}.}

\section{Algorithm}
This section explains how to use the RGP to learn heuristic functions from the benchmark dataset. Then, the learned G-heuristics are then applied in the GIT* to achieve fast and high-quality path planning. Finally, we prove that GIT* guarantees probabilistic completeness and asymptotic optimality.

\subsection{Notation}
\label{notation}
The state space of the planning problem is denoted by $X \subseteq \mathbb{R}^n$, where $n \in \mathbb{N}$. The start point is represented by $\mathbf{x}_{\text{start}} \in X$, and the goals are denoted by $X_{\text{goal}} \subset X$. The sampled states are denoted by $X_{\text{sampled}}$. The forward and reverse search trees are represented by $\mathcal{T_F} = (V_\mathcal{F}, E_\mathcal{F})$ and $\mathcal{T_R} = (V_\mathcal{R}, E_\mathcal{R})$, respectively. The vertices in these trees, denoted by $V_\mathcal{F}$ and $V_\mathcal{R}$, correspond to valid states. The edges in the forward tree, $E_\mathcal{F} \subseteq V_\mathcal{F} \times V_\mathcal{F}$, represent valid connections between states, while the edges in the reverse tree, $E_\mathcal{R} \subseteq V_\mathcal{R} \times V_\mathcal{R}$, may traverse invalid regions of the problem domain. An edge comprises a source state, $\mathbf{x}_s$, and a target state, $\mathbf{x}_t$, denoted as $(\mathbf{x}_s, \mathbf{x}_t)$. The true connection cost between two states in \textit{configuration space} (\textit{$\mathcal{C}$-space}) is represented by the function $c: X \times X \rightarrow [0, \infty)$.

Let $A$ be a set and let $B, C$ be subsets of $A$. The notation $B \stackrel{+}{\leftarrow} C$ is used to denote $B \leftarrow B \cup C$ and $B \stackrel{-}{\leftarrow} C$ is used to denote $B \leftarrow B \setminus C$.

\textit{GIT*-specific Notation:}
{Let $\Theta$ be the space of all path planning problems and $\Xi$ be the space of all path planning algorithms. The dataset consisting of \(k\) path planning problems is represented as \(D_{\text{benchmark}}^k = \{\theta_1, \theta_2, \ldots, \theta_k\}\). The function $\Phi: \Xi \times \Theta \to \{(m_1, v_1), (m_2, v_2), \ldots, (m_k, v_k)\}$ quantifies the expected performance of running an algorithm $\xi \in \Xi$ on a path planning problem $\theta \in \Theta$ one hundred times, with the performance measured by $k$ distinct indicators. The elements $(m_i, v_i)$ belong to a set $M \times \mathcal{V}$, where $M$ is the set of all possible metrics and $\mathcal{V}$ is the set of all possible values.}

{In the genetic programming process, an individual is denoted as $\psi \in \mathcal{E}$. The individual corresponding to the heuristic function of EIT* is defined as \(\psi_{EIT^*}\). The individuals generated in the same iteration form a population \(\mathcal{P}\), with size denoted as \(\mathcal{O}\). The probabilities of mutation and crossover, the two types of genetic operations, are denoted as $p_m$ and $p_c$ respectively. We define the fitness loss function $\phi: \mathcal{E} \to [0, \infty)$, which quantitatively evaluates the performance of individuals on the dataset \(D_{\text{benchmark}}^k\). The evaluated fitness value of an individual is denoted as $\rho_\psi := \phi(\psi)$. The algorithm obtained by substituting the heuristic function in EIT* with the individual \(\psi_i\) is denoted by \(GIT^*_{\psi_i}\). The reward function to assess the improvement of algorithm performance is denoted as $\chi: M \times \mathcal{V} \rightarrow [0, \infty)$. The function \(U: X \to [0, \infty)\) provides the magnitude of potential energy of a state in APF.}

\subsection{Reinforced Genetic Programming (RGP)}\label{sec: RGP}
This subsection introduces RGP and its adaptation to improve the heuristic function in sampling-based path planning. RGP uses a reward function to evaluate candidate models on unlabeled data, enabling model evolution.

As shown in Fig.~\ref{fig:GenerationIndividual}, RGP continues the traditional genetic programming's (GP) iterative evolutionary process. Initially, a primitive set is established to generate individuals and populations in the evolutionary cycle. This set includes essential components for individual generations. An algorithm outlines the rules for assembling individuals from these components. Multiple individuals created using the primitive set form a population of candidate solutions. Each individual $\psi_i$ represents a heuristic function and corresponds to a new algorithm $GIT^*_{\psi_i}$. The performance of this new algorithm is assessed using a Reinforced Fitness Evaluation Function, which compares the fitness of $GIT^*_{\psi_i}$ with the baseline (EIT*) algorithm.

Based on fitness values, exceptional individuals from the previous generation are chosen for the next, preserving superior genetic segments and removing inferior ones. This iterative process involves selecting parents, performing crossover and mutation to introduce new genetic segments, and evaluating the new population's fitness. Crossover mixes genetic material between parents, creating offspring with diverse traits, while mutation introduces random changes for unique variations. The process continues iteratively until a termination condition, such as a specific number of generations or satisfactory fitness, is met. The best-performing individual $\psi^*$ is then selected as the genotype-generative heuristic function for GIT*. The pseudocode for this process is illustrated in Alg.~\ref{alg:RGP}.

Unlike traditional GP, RGP employs a reward function to assess the fitness of individuals, known as the Reinforced Fitness Evaluation Function. In traditional GP, the dataset \(D\) comprises input data \(x_i \in X\) and corresponding label data \(l_i \in L\). The objective is for each individual's model, \(\psi_i\), to simulate the mapping from inputs to labels, \(\phi: X \rightarrow L\), minimizing the discrepancy between predicted outputs and actual labels to optimize model performance.
\begin{figure*}[t!] 
    \centering 
    \includegraphics[width=0.8\textwidth]{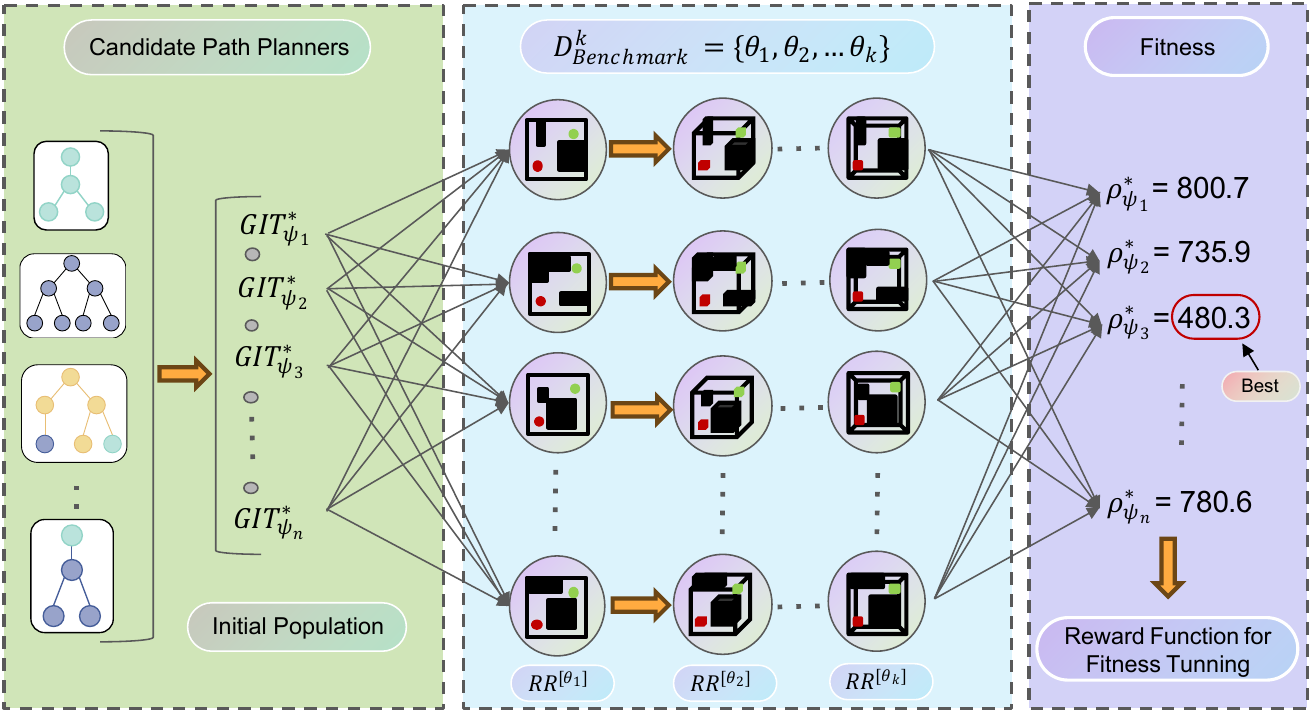} 
    \caption{llustration of the evaluation and fitness assignment process for individuals in the RGP. Each individual $\psi_i$ represents an expression that serves as a heuristic to guide the search, forming a new planner $GIT^*_{\psi_i}$. These planners are benchmarked across multi-dimensional and multi-scenario problems. They are scored based on a designed reward function, and the resulting score is the fitness value of the individual.}
    \label{fig:Genetic_operations} 
    \vspace{-1.7em} 
\end{figure*}
\begin{algorithm}[t!]
\label{alg:RGP}
\caption{Reinforced Genetic Programming (RGP)}
\SetKwInOut{Input}{Input}
\SetKwInOut{Output}{Output}
\SetKwFunction{initializePopulation}{initializePopulation}
\SetKwFunction{evaluateFitness}{evaluateFitness}
\SetKwFunction{terminateCondition}{terminateCondition}
\SetKwFunction{select}{select}
\SetKwFunction{crossover}{crossover}
\SetKwFunction{mutate}{mutate}
\SetKwFunction{bestIndividual}{bestIndividual}
\SetKwFunction{converged}{converged}

\DontPrintSemicolon

\Input{{Population size $\mathcal{O}$}, mutation rate $p_m$, crossover rate $p_c$}
\Output{Best found individual $\psi^*$}
$\mathcal{P} \gets \initializePopulation({\mathcal{O}})$

\While{\textbf{not} $\terminateCondition$}{
${\rho_\psi} \gets \evaluateFitness(\mathcal{P})$\\
$\mathcal{P}_{new} \gets \emptyset$\tcp{Initialize new population}

\For{$i \gets 1$ \textbf{to} {$\mathcal{O}$}}{
$\psi_{parent1} \gets \select(\mathcal{P}, {\rho_\psi})$\\
$\psi_{parent2} \gets \select(\mathcal{P}, {\rho_\psi})$\\

$\psi_{child} \gets \crossover(\psi_{parent1}, \psi_{parent2}, p_c)$

\If{\textbf{random}() $< p_m$}{
$\psi_{child} \gets \mutate(\psi_{child})$
}
$\mathcal{P}_{new} \stackrel{+}{\leftarrow} child$
}
$\mathcal{P} \gets \mathcal{P}_{new}$\\

$\psi^* \gets \bestIndividual(\mathcal{P})$
}
\Return{$\psi^*$}
\end{algorithm}

However, in our path planning problem, only the environment and problem description are provided as input data without any labels. We adopted an incentive-based approach to evaluate an individual’s fitness using our unlabeled dataset \(D_{\text{benchmark}}\). The objective is to identify an individual \(\psi^*\) from the set \(\mathcal{E}\) to replace the heuristic of EIT* and maximize the performance improvement over the EIT* baseline, as measured by the loss function \(\phi\). This is intended to optimize the effectiveness of the algorithm by adjusting its G-heuristic:
\begin{align}
\psi^* &= \arg\min_{\psi_i \in \mathcal{E}} \phi(\psi_i, D_{\text{benchmark}}), \\
\phi(\psi_i, D_{\text{benchmark}}) &= \chi(\Phi(\psi_i, D_{\text{benchmark}}), \Phi(\psi_{\text{EIT*}}, D_{\text{benchmark}})).
\end{align}

For problem description input \(\theta_i\), we use the existing algorithm EIT* performance as a control group, assessed over a set number of trials. We then compare this with the performance of a new algorithm \(GIT^*_{\psi_i}\), generated by replacing EIT*'s heuristic function with individual \(\psi_i\), tested under identical conditions. If \(GIT^*_{\psi_i}\) outperforms EIT* on any performance metric, the reward function \(\chi\) decreases the fitness score proportionally to the degree of improvement. Conversely, if \(GIT^*_{\psi_i}\) performs worse than EIT* on any metric, \(\chi\) increases the fitness score accordingly. This method ensures that a lower fitness score indicates the superior performance of an individual compared to EIT* within the dataset \(D_{\text{benchmark}}\).

To illustrate how the fitness of an individual $\rho_{\psi_i}$ is assessed, consider the following example. Table \ref{tab:performance_results} presents the performance results of EIT* and $GIT^*_{\psi_i}$. These two algorithms were tested 100 times on the Random Rectangle problems across different dimensions, with time limits for each run (unsuccessful runs were considered as infinite costs). Ten critical metrics were evaluated, reflecting the algorithm's performance in terms of \textit{time} to find the initial solution, the \textit{cost of the initial solution}, the \textit{cost of the optimal solution} within the time limit, and the final success rate over 100 runs of finding solutions.

When assessing the fitness $\rho$ of an individual $\psi_i$, these metrics must be taken into consideration, and the corresponding weights for each metric should be set according to the specific application context. Below, we present the \textbf{reward function system} and rules used in our subsequent experiments:

\textbf{1) Initial score:}  The initial score for an individual is 800, ensuring the final computed fitness is greater than 0.
    
\textbf{2) Weights of metrics:} To determine the specific weights \(w[m^i]\) for each metric \(m^i\), the weighting depends on the specific application scenario and requirements. The weights for the metrics are provided in the \textbf{weights} row of Table~\ref{tab:performance_results}.

\textbf{3) Base score for each metric:} For each metric \(m^i\), a base score \(s_\text{base}\) is assigned based on whether \(GIT^*_{\psi_i}\) outperforms \(EIT^*\) and the magnitude of the difference. First, it is assessed whether \(GIT^*_{\psi_i}\)'s value \(v^i_{GIT^*_{\psi_i}}\) outperforms \(EIT^*\)'s value \(v^i_{EIT^*}\). If \(GIT^*_{\psi_i}\) is superior, a fixed score \(\delta\) is subtracted; otherwise, it is added. To quantify the degree of superiority, this score is multiplied by a coefficient \(\alpha\), calculated as the ratio of the difference between \(v^i_{GIT^*_{\psi_i}}\) and \(v^i_{EIT^*}\):
\begin{equation}
    \alpha = \frac{v^i_{GIT^*_{\psi_i}} - v^i_{EIT^*}}{v^i_{EIT^*}},
\end{equation}

The base score for the metric \(m^i\) is then:
\begin{equation}
    s_\text{base}[m^i] = \delta + \delta \times \alpha ,
\end{equation}

\textbf{4) Handling infinity as a special case:} Some metrics may be infinite if solutions are not found in time. If both EIT* and $GIT^*_{\psi_i}$ record infinity for a metric, \( s_\text{base} \) is set to 0. If only one does, \( s_\text{base} \) is \( 2 \times \delta \).

\textbf{5) Bonus for significant success rate enhancement:} Given the importance of the success rate, substantial differences between $GIT^*_{\psi_i}$ and EIT* in this metric should impact the overall fitness evaluation. If the difference $v^{\text{success}}_{GIT^*_{\psi_i}} - v^{\text{success}}_{EIT^*}$ exceeds 5\% but is less than 15\%, a bonus \( s_{\text{bonus}}[m^{\text{success}}] = \delta \) is applied. For differences exceeding 15\%, \( s_{\text{bonus}}[m^{\text{success}}] = 2 \times \delta \).

\textbf{6) Calculation of total score:} The total score run on path planning problem $\theta$ is equal to the sum of all metrics' basic scores and bonuses, each multiplied by their respective weights. The total scure is expressed as:
    \begin{equation}
    s^{\theta}_{\text{total}} = \sum_{i=1}^n (s_{\text{base}}[m^i] + s_{\text{bonus}}[m^i]) \cdot w[m^i],
    \end{equation}
\begin{table*}[ht]
    \centering
    \caption{Example performance results for \textbf{randomly generated} environment}
    \label{tab:performance_results}
    \begin{tabularx}{0.7\textwidth}{ccccccccccc}
        \toprule
         & \(t_\mathrm{init}^\mathrm{min}\) & \textcolor{brown}{\(t_\mathrm{init}^\mathrm{med}\)} & \(t_\mathrm{init}^\mathrm{max}\) & \(c_\mathrm{init}^\mathrm{min}\) & 
         \textcolor{brown}{\(c_\mathrm{init}^\mathrm{med}\)} & \(c_\mathrm{init}^\mathrm{max}\) & \(c_\mathrm{final}^\mathrm{min}\) & 
         \textcolor{brown}{\(c_\mathrm{final}^\mathrm{med}\)} & \(c_\mathrm{final}^\mathrm{max}\) & 
         \textcolor{brown}{Success} \\[0.5em]

        \textcolor{teal}{Weights $w[m^i]$} & \textcolor{teal}{1.0} & \textcolor{teal}{3.5} & \textcolor{teal}{0.5} & \textcolor{teal}{1.0} & \textcolor{teal}{2.5} & \textcolor{teal}{1.0} & \textcolor{teal}{1.0} & \textcolor{teal}{2.5} & \textcolor{teal}{1.0} & \textcolor{teal}{3.0} \\
                \midrule
        EIT* & 0.19 & $\infty$ & $\infty$ & 2.5 & $\infty$ & $\infty$ & 2.5 & $\infty$ & $\infty$ & 0.48 \\[0.5em]
        GIT$^*_{\psi_i}$ & \bfseries 0.16 & \bfseries 0.39 & $\infty$ & \bfseries 2.34 & \bfseries 5.05 & $\infty$ & \bfseries 2.33 & \bfseries 3.53 & $\infty$ & \bfseries 0.72 \\[0.5em]
        \bottomrule
    \end{tabularx}
    \vspace{-1.7em} 
\end{table*}

The above rules evaluate an individual's total score within a specific problem context. To measure generalizability, we use randomly generated problem descriptions as a dataset \(D_{\text{benchmark}}\). The average total scores within this dataset are included in the fitness calculation. To ensure stability across problems, we include the variance of total scores. Lastly, we consider the number of nodes as a complexity measure to avoid overfitting from complex expressions. The final fitness calculation formula is as follows:
\begin{equation}
\rho_\psi = \overline{s_{\text{total}}} + c_1 \sigma^2_{s_{\text{total}}} + c_2 |\psi|.
\end{equation}
where $\psi$ denotes the individual. $\overline{s_{\text{total}}}$  represents the mean of the total scores across each problem definition in the benchmark. $\sigma^2_{s_{\text{total}}}$ is the variance of the total scores, multiplied by the coefficient $c_1$. $|\psi|$ signifies the size of the individual $\psi$, multiplied by the coefficient $c_2$.

During practical training, techniques can reduce unnecessary computations. A segmented system can evaluate an individual without testing the complete benchmark. The benchmark's scenarios are divided into segments with increasing difficulty. If the fitness in the first $i$ segments is significantly lower than a baseline, it indicates the algorithm performs worse than the expected ideal threshold (EIT*). Consequently, the fitness score can be directly assessed and recorded as $L_\psi$ without testing the entire benchmark, as such an individual is likely to be quickly eliminated in the evolutionary process.
\begin{algorithm}[t]
\caption{Genetic Informed Trees (GIT*)}
\SetKwInOut{Input}{Input}
\SetKwInOut{Output}{Output}
\SetKwFunction{sample}{sample}
\SetKwFunction{bestKey}{bestKey}
\SetKwFunction{reverseSearch}{reverseSearch}
\SetKwFunction{updateReverseSearch}{updateReverseSearch}
\SetKwFunction{forwardSearch}{forwardSearch}
\SetKwFunction{couldImproveForwardSearch}{couldImproveForwardSearch}
\SetKwFunction{pathFound}{pathFound}
\SetKwFunction{lazyCheck}{lazyCheck}
\SetKwFunction{terminateCondition}{terminateCondition}
\SetKwFunction{prune}{prune}

\DontPrintSemicolon

\Input{$\text{Start point}~\mathbf{x}_{\text{start}}$, \text{goal region}~$X_{\text{goal}}$, \text{best individual}~\(\psi^*\), \text{optimal inflation/truncation factors}~$\varepsilon^*_{\text{infl}}$, $\varepsilon^*_{\text{trunc}}$}
\Output{$\text{Feasible path}~\mathcal{T_F}$}
$X_{\textit{sampled}} \gets \{\mathbf{x}_{\text{goal}}\}$, $E_\mathcal{F} \gets \emptyset$, $\mathcal{T_F} = (V_\mathcal{F}, E_\mathcal{F})$\;

$\mathrm{key}_{\mathcal{R}}^{\mathrm{GIT}^*} \gets \bestKey(\psi^*)$\\
\While{\textbf{not} $\terminateCondition()$}{
    $X_{\textit{sampled}} \stackrel{+}{\leftarrow} \sample()$\\
    $\mathcal{T_R}\gets\reverseSearch(\mathrm{key}_{\mathcal{R}}^{\mathrm{GIT}^*},\textcolor{purple}{\varepsilon^*_{\text{infl}}}, \textcolor{purple}{\varepsilon^*_{\text{trunc}}})$\\
    \While{$\couldImproveForwardSearch(\mathcal{T_R})$}
    {   
        $E_\mathcal{F} \gets\forwardSearch(\mathcal{T_R}, \textcolor{purple}{\varepsilon^*_{\text{trunc}}})$\\

        \eIf{$\pathFound(E_\mathcal{F})$}{
        $\mathcal{T_F}\gets\mathcal{T_R}$\\

    }{$\updateReverseSearch()$\\}}
        $\prune(\mathcal{T_F})$\\
        }{\Return {$\mathcal{T_F}$}

}
\end{algorithm}
\begin{algorithm}[t]
\caption{GIT*: Potential Energy}
\label{alg:potentialEnergy}
\DontPrintSemicolon
\SetKwFunction{calcAttractiveEnergy}{calcAttractiveEnergy}
\SetKwFunction{calcRepulsiveEnergy}{calcRepulsiveEnergy}
\SetKwIF{If}{ElseIf}{Else}{if}{}{else if}{else}{end if}%
$U[\mathbf{x}] \leftarrow \emptyset$\;
$U_{\text{attr}}[\mathbf{x}] \leftarrow \calcAttractiveEnergy(\mathbf{x}_{\text{start}}, \mathbf{x})$
\tcp{Equation~\ref{Uattr}}
\ForEach{$\mathbf{x}_{\text{invalid}} \in X_{\text{invalid}}$}{
    $U_{\text{rep}}[\mathbf{x}]  \mathrel{+}= \calcRepulsiveEnergy(\mathbf{x}_{\textit{invalid}}, \mathbf{x})$\\
    \tcp{Equation~\ref{Urep}}
}
$U[\mathbf{x}] \leftarrow U_{\text{rep}}[\mathbf{x}] + U_{\text{attr}}[\mathbf{x}]$\;
\Return{$U[\mathbf{x}]$}
\end{algorithm}
\subsection{Genetic Informed Trees (GIT*)}\label{3.4GIT*}

In Section~\ref{sec: RGP}, we use the RGP to evaluate the best individual $\psi^*$ of the generated population. In this subsection, the evaluated best individual $\psi^*$ is utilized in the GIT* to guide robot path planning, allowing the robot to rapidly converge on the initial solution while maintaining path quality.

Problem-specific information falls into three categories: search tree information $g(\mathbf{x})$, heuristic information $\hat{h}(\mathbf{x})$, and environmental information (e.g., dimensionality $D(\theta)$ and obstacle details). GIT* uses the RGP to generate evolving individuals that combine these information types into complex expressions. These expressions are integrated into the EIT* heuristic function to form new algorithms, $GIT^*_{\psi}$, with the optimal GIT* algorithm being selected based on performance:
\begin{align}
\psi^* &:= \arg\min_{\psi_i} \rho_{\psi_i},
\\
GIT^* &:= GIT^*_{\psi^*},
\end{align}

When GIT* trains its heuristic function using RGP, information is stored in the primitive set to generate individuals. The search tree-related information includes $g(\mathbf{x}_s)$, while prior heuristic information includes $\hat{h}(\mathbf{x}_t)$ and $\hat{c}(\mathbf{x}_s,\mathbf{x}_t)$. \(\bar{e}(\mathbf{x}_{\mathrm{s}})\) estimates the effort to find and validate a path from \(\mathbf{x}_{\mathrm{s}}\) to the goal, whereas \(\bar{e}(\mathbf{x}_{\mathrm{s}}, \mathbf{x}_{\mathrm{t}})\) estimates the computational effort required to find and validate a path between states, while \(\bar{d}(\mathbf{x}_{\mathrm{t}})\) estimates the effort from \(\mathbf{x}_{\mathrm{t}}\) to the start. Environmental information comprises not only $D(\theta)$ but also two variables that record information about obstacles and the dynamic importance of states. According to the GA model, after natural selection, the \textbf{winner G-heuristic} function generated by RGP can be equivalently represented by $\mathrm{key}_{\mathcal{R}}^{\mathrm{GIT}^*}$, which extracts the next edge from the reverse queue:
\begin{equation}\label{equ:key}
\mathrm{key}_{\mathcal{R}}^{\mathrm{GIT}^*}\left( \mathbf{x}_{\mathrm{s}},\mathbf{x}_{\mathrm{t}} \right) :=
\begin{cases}
\left(\widehat{g}\left( \mathbf{x}_{\mathrm{t}} \right) - \pi\right) \times \frac{\log\left(1 + \left| U[\mathbf{x}_t] - U[\mathbf{x}_s]\right|\right)}{1 + w_{\textit{dyn}}[\mathbf{x}_t]},\\
\sqrt{\bar{e}(\mathbf{x}_{\mathrm{s}}) + \bar{e}(\mathbf{x}_{\mathrm{s}}, \mathbf{x}_{\mathrm{t}})} \times \log(\bar{d}(\mathbf{x}_{\mathrm{t}})).
\end{cases}
\end{equation}
where $U[\mathbf{x}_t]$ refers to the potential energy of the current state in an artificial potential field, and $w_{\text{dym}}$ refers to dynamic importance, represented by the number of times the current state has been visited. The following will detail how these variables are obtained.

\subsubsection{\textbf{Potential field variable} $U[\mathbf{x}_t]$}\label{3.4.1Obstacle}
Understanding obstacle characteristics like shapes, numbers, and locations is crucial for guiding the search tree to either circumvent obstacles for quicker solutions or approach them to reduce costs. However, these characteristics are often unknown beforehand.

{GIT* approximates the environment by sampling points in the \textit{$\mathcal{C}$-space} to acquire information about obstacles, denoted as ${X}_\text{obs}$. Those randomly sampled points undergo a validity check (e.g., collision detection) to determine if they are inside obstacles. Invalid points, denoted as $\mathbf{x}_{\text{invalid}}$, indicate locations within obstacles, gradually outlining their shapes and locations as sampling increases. GIT* also employs the APF method to conceptualize the navigation space as a force field where obstacles generate repulsive forces, and targets generate attractive forces (Alg.~\ref{alg:potentialEnergy}, line 2). $\mathbf{x}_{\text{invalid}}$ and $\mathbf{x}_{\text{goal}}$ generate repulsive and attractive forces with target state $\mathbf{x}_{\text{t}}$, respectively, and the potential field is dynamically adjusted based on the obstacle data. The calculated data is then utilized in the primitive set as candidates for RGP to generate G-heuristic individuals.}

\begin{itemize}
    \item \textbf{Repulsive 
    force}: Generated around invalid samples, these forces prevent entry into these areas. The magnitude of the repulsive force is:
    \begin{equation}
    F_{\text{rep}}(q) := \left\{
    \begin{array}{ll}
        -\frac{k_r \cdot q \cdot q_{\text{obs}}}{r^2} & \text{if } r \leq \rho_0 \\
        0 & \text{otherwise}
    \end{array},
    \right.    
    \end{equation}
    where \( k_r \) is a proportionality constant, \( q \) is the charge equivalent of the path planner, \( q_{\text{obs}} \) is the charge equivalent of the obstacle, \( r \) is the distance between the path planner and the obstacle, and \( \rho_0 \) is the threshold distance beyond which the force is not exerted.

    \item \textbf{Repulsive potential energy}: The potential energy is:
    \begin{equation}
    \label{Urep}
    U_{\text{rep}}(q) := \left\{
    \begin{array}{ll}
        -\frac{k_r \cdot q \cdot q_{\text{obs}}}{r} & \text{if } r \leq \rho_0 \\
        0 & \text{otherwise}
    \end{array},
    \right.    
    \end{equation}

    \item \textbf{Attractive force}: Produced by the target, these forces guide the path planner towards the target, navigating around repulsive regions. The magnitude of the attractive force is:
    \begin{equation}
    F_{\text{attr}}(q):= \frac{k_a \cdot q \cdot q_{\text{goal}}}{r^2}    ,
    \end{equation}
    where \( k_a \) is another proportionality constant, \( q \) is the charge equivalent of the path planner, \( q_{\text{goal}} \) is the charge equivalent of the target, and \( r \) is the distance between the path planner and the target.

    \item \textbf{Attractive potential energy}: The potential energy is:
    \begin{equation}
    \label{Uattr}
    U_{\text{attr}}(q) := \frac{k_a \cdot q \cdot q_{\text{goal}}}{r}.
    \end{equation}
\end{itemize}

The potential energy in the APF can be calculated using these formulae, recording information about obstacles and incorporating it into the primitive set to construct the heuristic function. As potential energy increases, indicating proximity to obstacles, the heuristic function's value increases, reducing the likelihood of state selection. When $U[\mathbf{x}_t]$ is high, indicating frequent visits, the heuristic function's value decreases, increasing the likelihood of exploration. As $\widehat{g}\left( \mathbf{x}_{\mathrm{t}} \right)$ increases, indicating greater distance from the start, the heuristic function's value increases, making the node less likely to be searched.

\subsubsection{\textbf{Dynamic importance variable} $w_{\textit{dyn}}
[\mathbf{x}_t]$}\label{3.4.2Dynamic}
\begin{algorithm}[t]
\label{alg: Dynamic}
\caption{GIT*: Dynamic Importance}
\DontPrintSemicolon
\SetKwFunction{neighbors}{neighbors}
\SetKwFunction{inReverseTree}{inReverseTree}
\SetKwIF{If}{ElseIf}{Else}{if}{}{else if}{else}{end if}%
$w_{\textit{dyn}}
[\mathbf{x}_t] \leftarrow \emptyset$\;

\ForEach{$\mathbf{x}_{\text{neighbor}} \in \neighbors(\mathbf{x}_t) $}{
    \If{$\inReverseTree(\mathbf{x}_{\text{neighbor}}, \mathbf{x}_t)$}{
        $w_{\textit{dyn}}
[\mathbf{x}_t] \leftarrow w_{\textit{dyn}}
[\mathbf{x}_t] + 1$\;
    }
}

\Return{$w_{\textit{dyn}}
[\mathbf{x}_t]$}
\end{algorithm}
\begin{algorithm}[t]
\caption{GIT*: Nearest Neighbors}
\label{alg: neighbor}
\DontPrintSemicolon
\SetKwFunction{parent}{parent}
\SetKwFunction{children}{children}
\SetKwFunction{nearest}{nearest}
\SetKwIF{If}{ElseIf}{Else}{if}{}{else if}{else}{end if}%

$X_\textit{neighbors}\leftarrow \nearest(\mathbf{x}_t)$\;
$X_\textit{neighbors}\stackrel{+}{\leftarrow} \{\parent(\mathbf{x}_t)\cup\children(\mathbf{x}_t)\setminus X_\textit{neighbors}\}$\;
$X_\textit{neighbors}\stackrel{-}{\leftarrow} \{\mathbf{x}_s \in X_\textit{neighbors} \arrowvert (\mathbf{x}_s, \mathbf{x}_t) \in E_\textit{invalid}\}$\;

\Return{$X_\textit{neighbors}$}
\end{algorithm}

{In incremental asymptotically sampling-based planners like GIT*, certain sample points in \textit{$\mathcal{C}$-space} are frequently visited, often in nearest neighbor areas (e.g., path rewire) of path planning. These samples may lie along essential routes between start and end states, serve as conduits connecting regions, and could be located in narrow corridor areas. Thus, frequently visited samples in the free space prior to neighboring areas guide the search into explore-worthy regions, which improves search efficiency. GIT* tracks the number of visits to each sample point, capturing its dynamic importance (Alg.~\ref{alg: Dynamic} and~\ref{alg: neighbor}), and navigates to higher importance states. These strategies help GIT* search more efficiently during the path optimization phase. Similar to the APF discussed in Section~\ref{3.4.1Obstacle}, the number of visits (i.e., dynamic importance) to a state is included in RGP's primitive set to generate G-heuristics.}

Formally, the dynamic importance of a state $\mathbf{x}_t$, denoted as $w_{\textit{dyn}}[\mathbf{x}_t]$, is calculated as follows:
\begin{equation}
w_{\textit{dyn}}[\mathbf{x}_t] := \hspace{-0.8cm}\sum_{\mathbf{x}_{\text{neighbor}} \in X_\text{neighbors}(\mathbf{x}_t)} \hspace{-0.8cm}\mathbb{I}(\mathbf{x}_{\text{neighbor}} \in \mathcal{T_R}(X_\text{neighbors}(\mathbf{x}_t))),
\end{equation}
where $\mathbb{I}(\cdot)$ is the indicator function that equals 1 if the condition is true and 0 otherwise.

Each time a sample point appears in the nearest neighbors of the reverse tree queue, the dynamic importance of the corresponding state is incremented by 1, emphasizing frequently visited states for path optimization. Furthermore, the inflation factor speeds up the search by biasing the goal, resulting in rapid initial solutions. The truncation factor optimizes the search by stopping it when the solution quality is satisfied.

\subsubsection{\textbf{Inflation and truncation factor function}}
The traditional inflation and truncation factor update strategy is a user-adjustable parameter that can be tailored to specific application scenarios and requirements. However, this strategy lacks flexibility as it requires manual adjustments in each scenario to achieve optimal performance.
The updated function for the inflation factor derived from this training session is:
\begin{equation}
\varepsilon^*_{\text{infl}} = 1.0 + \frac{\log(D(\theta)) + \sqrt{D(\theta)}}{\sqrt{N_{\text{samples}}} + \log(N_{\text{samples}}) + 1},
\end{equation}
where $D(\theta)$ is the dimensionality of the path planning problem $\theta$, and $N_{\text{samples}}$ is the current number of samples taken.

As the problem's dimensionality increases, this expression's value also increases, biasing the search towards rapidly finding feasible solutions based on heuristics rather than ensuring the lowest cost solution. In higher-dimensional spaces, fewer obstacles relative to the overall space decrease the probability of blocking the path, enhancing the success rate and reducing the time to find initial solutions. As the number of samples increases, the value decreases, leading GIT* to focus on low-cost solutions after several sampling batches, aligning with practical requirements.

The updated function for the truncation factor is:
\begin{equation}
\varepsilon^*_{\text{trunc}} = 1.0 + \frac{3\pi}{N_{\text{samples}}},
\end{equation}
where $N_{\text{samples}}$ represents the current number of samples taken.

As $N_{\text{samples}}$ increases, the value decreases, indicating a tendency to exploit the current approximation rather than explore new ones. This is suitable for the later stages of the search when $N_{\text{samples}}$ is large.

\section{Analysis}

In this section, firstly, we provide the convergency analysis and asymptotical time to prove the feasibility of the proposed algorithm. In addition, we explain the reason that GIT* consumes less time complexity, and we also verify the advantage of reinforced genetic programming (RGP) from mathematics.

\subsection{Reinforced Genetic Programming Training Analysis}
\begin{figure}[h] 
    \centering 
    \includegraphics[width = 0.48\textwidth]{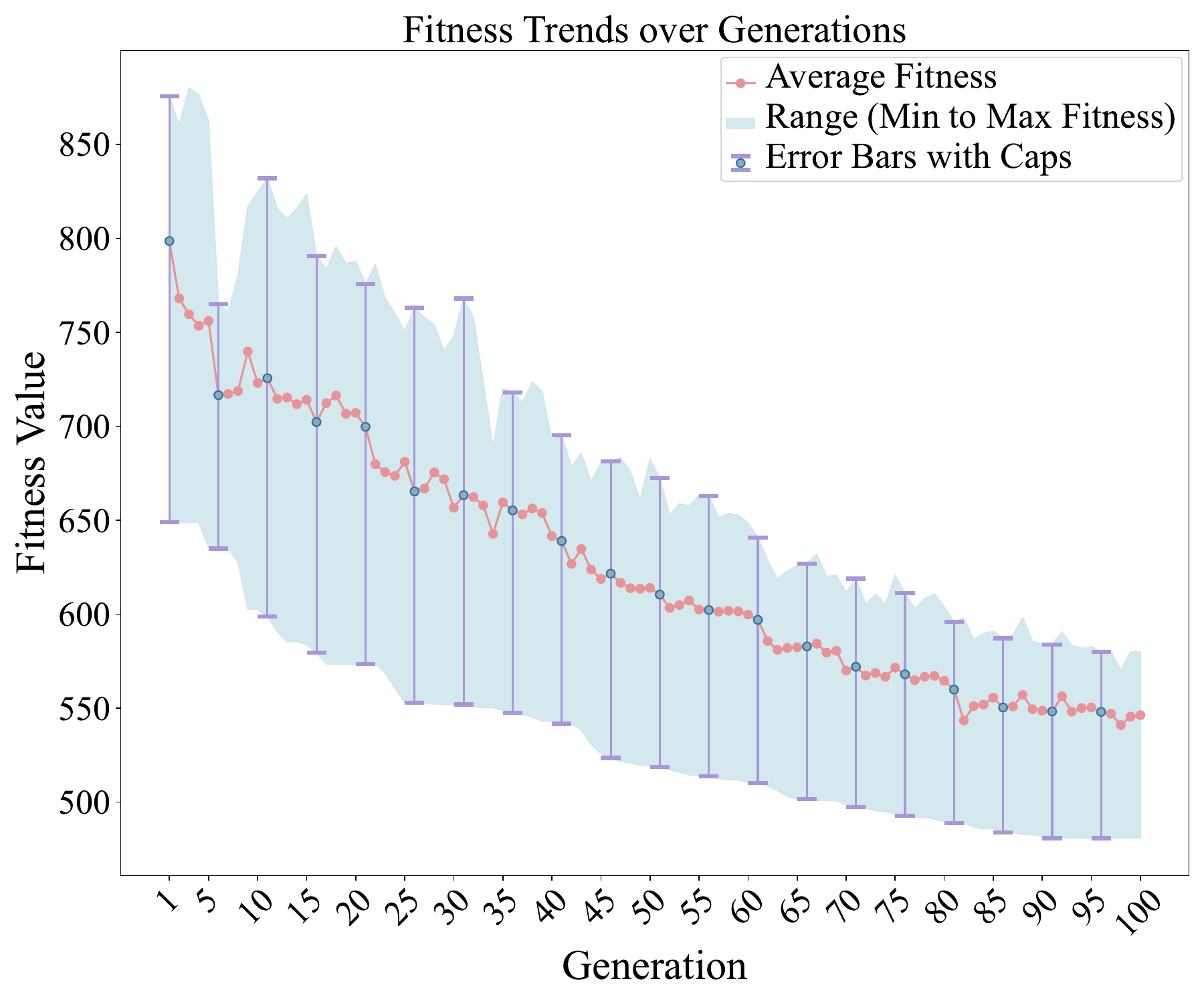} 
    \vspace{-0.8em} 
    \caption{{Evolution of fitness over generations during genetic programming training. The light purple error bar represents the maximum/minimum fitness value within each generation, while the light orange line depicts the average fitness value across individuals within each generation. The horizontal axis represents the generations, and the vertical axis represents the fitness values.}} 
    \label{fig:fitnessOverGeneration} 
    \vspace{-1.4em} 
\end{figure}

{Due to the randomness of the RGP algorithm and variability in training parameters, results from each RGP instance are unique. Practical applications need to consider specific objectives, use cases, datasets, and time constraints for parameter settings. Table~\ref{tab:params} details the chosen parameters: high population size enhances diversity but raises computational cost, 1500 size was chosen for optimal performance with our equipment; 100 generations provide a balance between solution quality and overfitting; a crossover rate of 0.8 promotes exploration without excessive disruption; a mutation rate of 0.1 maintains diversity and prevents premature convergence; a maximum tree depth of 4 avoids overfitting and underfitting; and a tournament size of 5 balances selection pressure and diversity. The fitness variation across generations is shown in Fig.~\ref{fig:fitnessOverGeneration}.}

\begin{table}[htb]
\centering
\caption{Parameter settings for the training}
\label{tab:params}
\resizebox{0.38\textwidth}{!}{%
\begin{tabular}{ll}
\toprule
Parameter Name    & Value or Description \\ 
\midrule
Population Size   & 1500 \\
Number of Generations        & 100\\
Selection Method            & Tournament Selection \\
Crossover Rate         & 0.8 \\
Mutation Rate     & 0.1 \\
Maximum Tree Depth & 4 \\
Tournament Size & 5 \\
Crossover Type & Subtree Crossover \\
Mutation Type & Point Mutation \\
\bottomrule
\end{tabular}
}
\end{table}

\subsection{{Proof of Convergence in Genetic Programming}}
Research has explored the convergence properties of genetic programming (GP) for symbolic regression~\cite{langdon2013foundations}. {The global optimum in symbolic regression problem refers to the best possible individual that achieves the minimum or maximum fitness of the objective function across the entire state space~\cite{horst1996global}.} Convergence to the global optimum implies generating solutions where the global optimum emerges as a limit. This study adopts a probabilistic interpretation. Rudolph \cite{rudolph1994convergence} modeled genetic programming using a Markov Chain framework and demonstrated convergence when the population retains the best solution. The natural selection, crossover, and mutation processes in GP mimic biological evolution.

Let \( \mathcal{P}(t) \) be the population at time \( t \), and \( \psi^* \) be the global optimum. GP maintains a diverse population \( \mathcal{P}(t) \) over generations to escape local optima:
\begin{equation}
   \mathcal{P}(t+1) := \textit{select}(\textit{crossover}(\textit{mutate}(\mathcal{P}(t)))),
\end{equation}
   This helps GP escape local optima, unlike greedy search methods, which may converge quickly to local optima.

Let \( Z_t \) denote a sequence of random variables representing the best fitness within a population at step \( t \). The convergence property of genetic programming, which preserves the best solution in the population, can then be formalized as:
\begin{equation}
\lim_{t \to \infty} \mathbb{P}(Z_t = \psi^*) := 1,
\end{equation}
where \(\psi^*\) represents the global optimum. This expression indicates that the probability of the best fitness \(Z_t\) equating to the global optimum \(\psi^*\) approaches unity as the number of iterations steps \(t\) approaches infinity.

Through mutation and crossover, GP maintains diversity and explores the search space effectively. Selection mechanisms favor individuals with higher fitness, leading to gradual improvement. Consequently, the probability of finding the global optimum \(\psi^*\) increases with each iteration.
\subsection{Probabilistic Completeness and Asymptotic Optimality}

Most informed tree-based path planning algorithms have been proven to be probabilistically complete and asymptotically optimal, and GIT* can also guarantee these two properties. GIT* utilizes uniform sampling strategies. As the number of iterations $n$ approaches infinity, the entire state space will be explored, satisfying the following equation:
\begin{equation}
\lim_{n \to \infty} \mathbb{P} (\{V_\mathcal{F}\cup V_\mathcal{R}\} \cap X_{\text{goal}}) \neq \emptyset) = 1, 
\end{equation}
which means that if there is a feasible path, it must be found by the GIT*. Therefore, the probabilistic completeness of the optimal path planner is guaranteed. 

The GIT* implements the same Choose Parent and Rewire strategies as the EIT*. It means that if the rewiring radius $r(q)$  in Choose Parent and Rewire processes satisfies:
\begin{equation}
\label{eqn:radius r}
    r(q) > \eta \left(2 \left(1 + \frac{1}{d}\right){\left(\frac{\lambda(X_{\hat{f}})}{\zeta_d}\right) \left( \frac{\log(q)}{q}\right)}\right)^{\frac{1}{d}},
\end{equation}
here, $q$ denotes the number of sampled states in the informed set, $\eta >$ 1 is a tuning parameter, $\lambda(\cdot)$ denotes the Lebesgue measure, and  $d$ is the dimensionality of the workspace, $\lambda(X_{\hat{f}}))$ is the Lebesgue measure of informed set $X_{\hat{f}}$ and $\zeta_d$ is the volume of unit ball in current workspace. In reference to Lemma 56, 71 and 72 in~\cite{karaman2011sampling}, the following equation holds:
\begin{equation}
\mathbb{P} (\limsup_{q \to \infty} \min_{\sigma\in\Sigma_q} \left\{ c(\sigma) \right\} = c^*) = 1,
\end{equation}
where $q$ is the number of samples, $\Sigma_q \subset \Sigma$ is the set of valid paths from the start to the goal found by the planner from those samples, $c: \Sigma \rightarrow [0, \infty)$ is the cost function, and $c^*$ is the optimal solution cost. It indicates that the GIT* can find an optimal path, if it exists, as the number of iterations go to infinity. Therefore, the asymptotic optimality is guaranteed.

\section{Experiments}\label{sec:Expri}
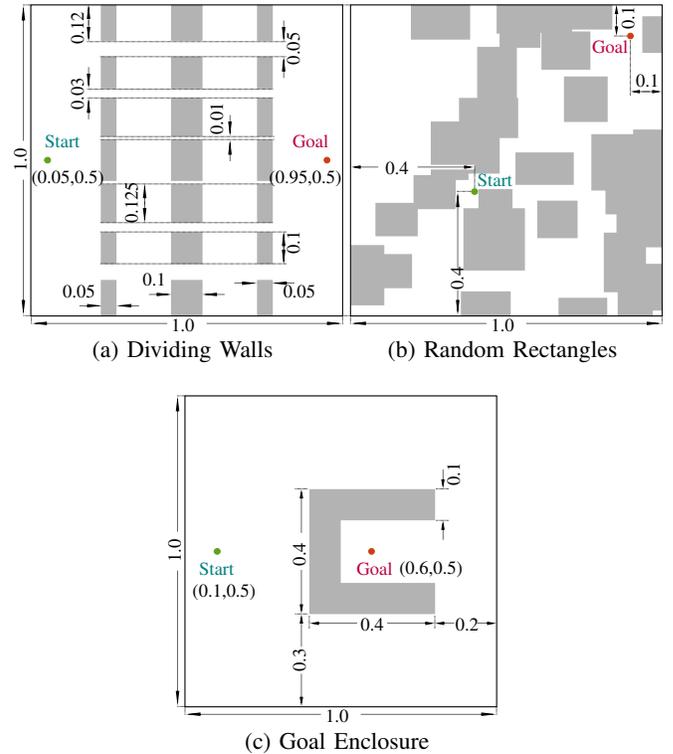
\begin{figure}[t!]
    \centering
    \begin{tikzpicture}
    \node[inner sep=0pt] (russell) at (-4.0,0.0)
    {\includegraphics[width=0.24\textwidth]{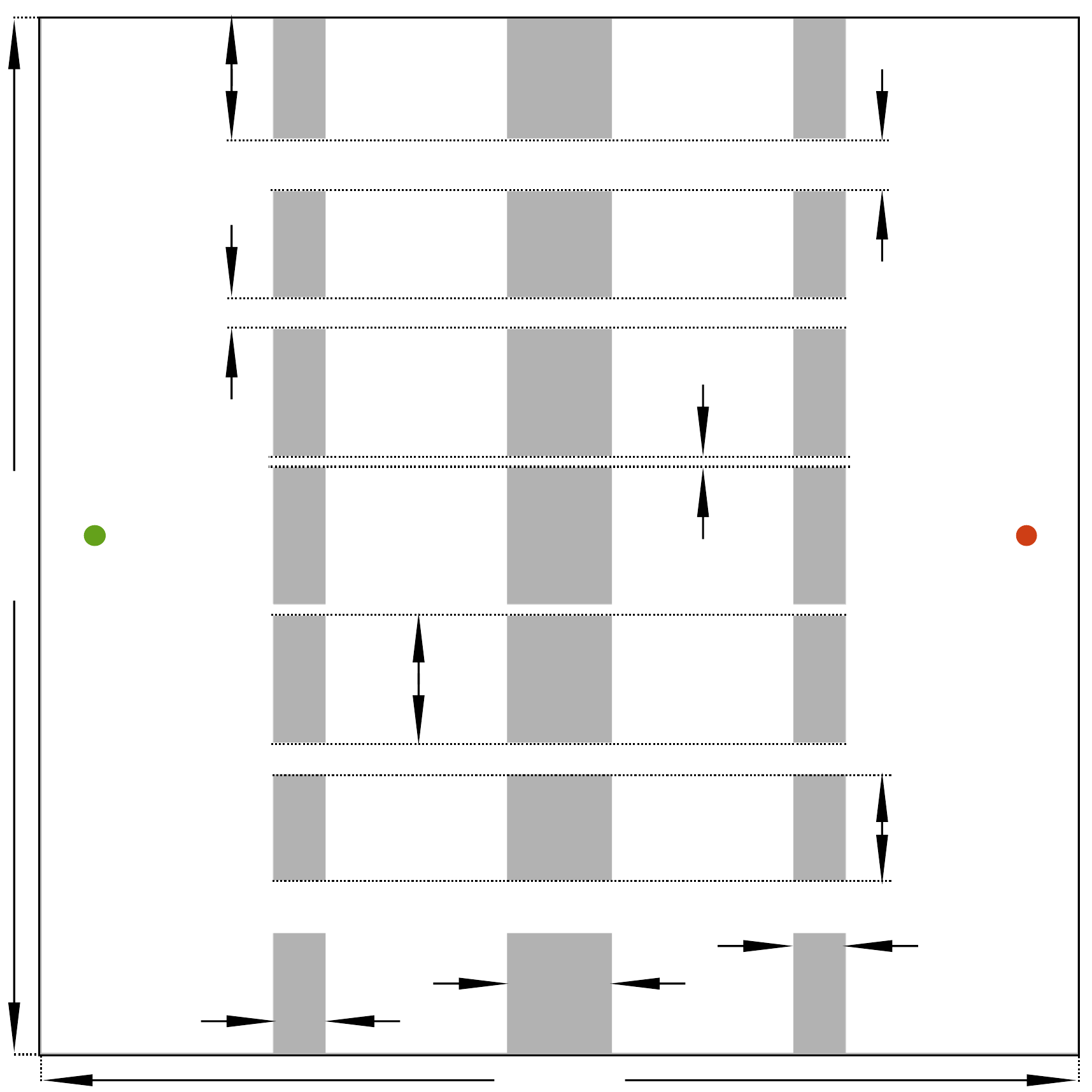}};
    \node[inner sep=0pt] (russell) at (0.25,0.0)
    {\includegraphics[width=0.24\textwidth]{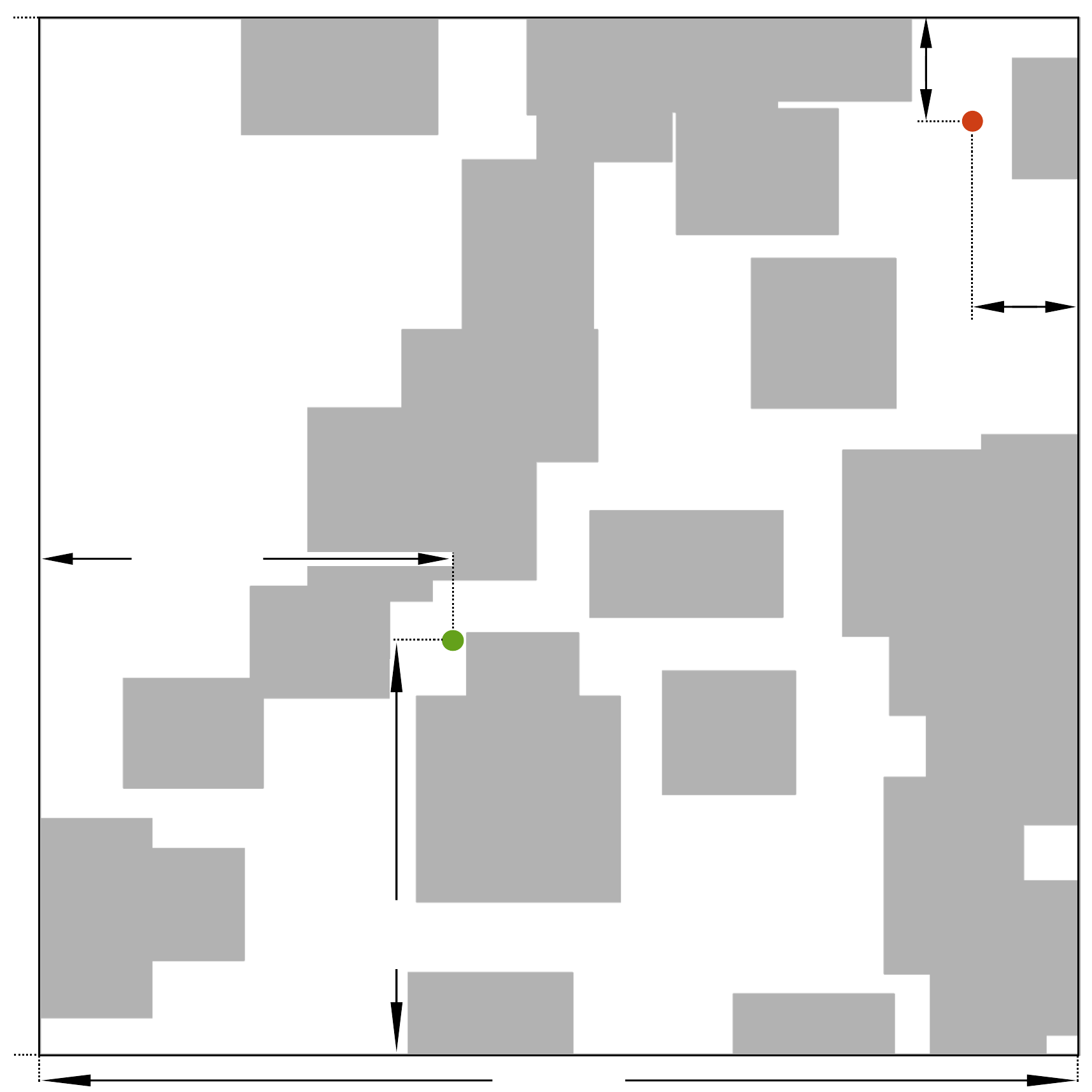}};
    \node[inner sep=0pt] (russell) at (-1.95,-5.2)
    {\includegraphics[width=0.24\textwidth]{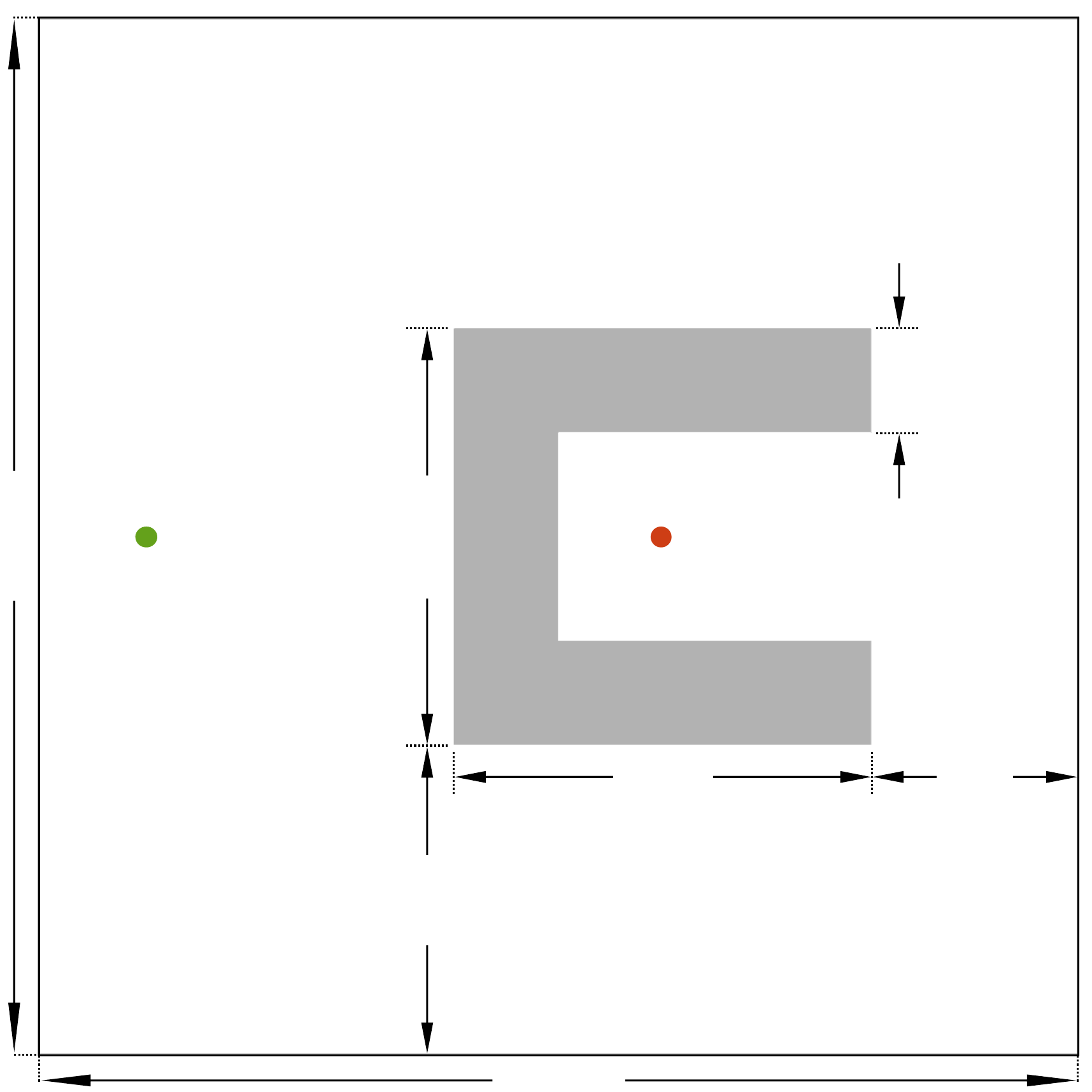}};
    \scriptsize

    \node [rotate=90] at (-5.4,0.93) {0.03};    
    \node [rotate=90] at (-4.68,-0.52) {0.125};    
    \node [rotate=90] at (-5.4,1.8) {0.12};
    \node [rotate=90] at (-2.53,-1.05) {0.1};
    \node [rotate=90] at (-3.55,0.6) {0.01};
    \node [rotate=90] at (-2.53,1.53) {0.05};
    \node at (-5.38,-1.7) {0.05};
    \node at (-4.38,-1.55) {0.1};
    \node at (-2.45,-1.7) {0.05};
    \node [rotate=90] at (-6.16,0.05) {1.0};
    \node at (-4.0,-2.15) {1.0};
    \node at (-5.53,-0.2) {(0.05,0.5)};
    \node at (-2.37,-0.2) {(0.95,0.5)};
    \node at (-5.6,0.3) {\color{teal} Start};
    \node at (-2.3,0.3) {\color{purple} Goal};

    \node at (-1.15,-0.06) {0.4};
    \node at (0.25,-2.15) {1.0};
    \node [rotate=90] at (-0.35,-1.53) {0.4};
    \node [rotate=90] at (1.92,1.94) {0.1};
    \node  at (2.18,1.1) {0.1};

    \node at (0.15,-0.22) {\color{teal} Start};
    \node at (1.68,1.55) {\color{purple} Goal};
    
    \node at (-1.48,-6.13) {0.4};
    \node at (-1.95,-7.35) {1.0};
    \node [rotate=90] at (-4.11,-5.15) {1.0};
    \node [rotate=90] at (-0.4,-4.13) {0.1};
    \node [rotate=90] at (-2.45,-5.2) {0.4};
    \node [rotate=90] at (-2.45,-6.61) {0.3};

    \node at (-0.22,-6.13) {0.2};
    \node at (-3.45,-5.7) {(0.1,0.5)};
    \node at (-0.7,-5.4) {(0.6,0.5)};

    \node at (-3.55,-5.4) {\color{teal} Start};
    \node at (-1.45,-5.4) {\color{purple} Goal};

    \node at (-4.0,-2.51) {\small (a) Dividing Walls};
    \node at (0.25,-2.51) {\small (b) Random Rectangles};
    \node at (-1.95,-7.71) {\small (c) Goal Enclosure};
    \end{tikzpicture}
    \caption{The 2D representation of the simulated planning problems in Section~\ref{sec:Expri}. The state space, denoted as $X \subset \mathbb{R}^n$, is constrained within a hypercube with one width for both problem instances. Specifically, we conducted ten distinct instantiations of the random rectangles experiment and the outcomes are showcased in Fig.~\ref{fig: result}.}
    \label{fig: testEnv}
    \vspace{-1.7em} 
\end{figure}
\begin{figure*}[t!]
    \centering
    \begin{tikzpicture}
    \node[inner sep=0pt] (russell) at (4.1,8)
    {\includegraphics[width=0.49\textwidth]{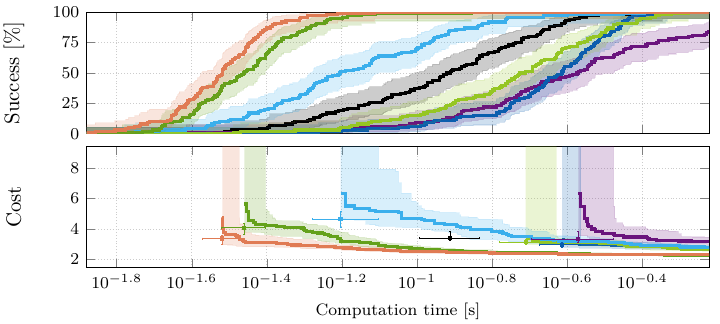}};
    \node[inner sep=0pt] (russell) at (4.1,3.5)
    {\includegraphics[width=0.49\textwidth]{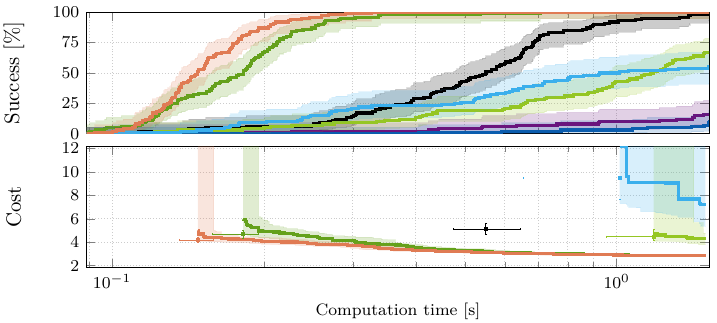}};
    \node[inner sep=0pt] (russell) at (4.1,-1)
    {\includegraphics[width=0.49\textwidth]{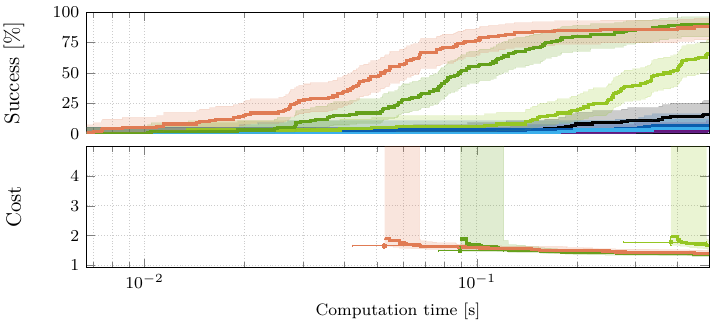}};

    \node[inner sep=0pt] (russell) at (-4.9,8)
    {\includegraphics[width=0.49\textwidth]{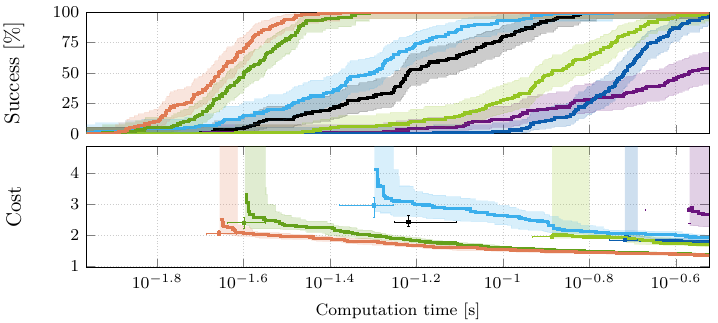}};  
    \node[inner sep=0pt] (russell) at (-4.93,3.5)
    {\includegraphics[width=0.489\textwidth]{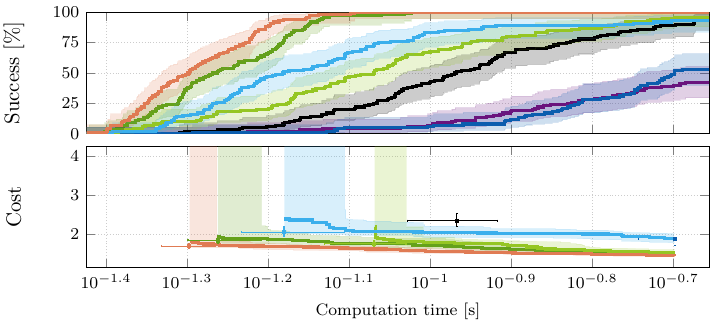}};
    \node[inner sep=0pt] (russell) at (-4.85,-1){\includegraphics[width=0.496\textwidth]{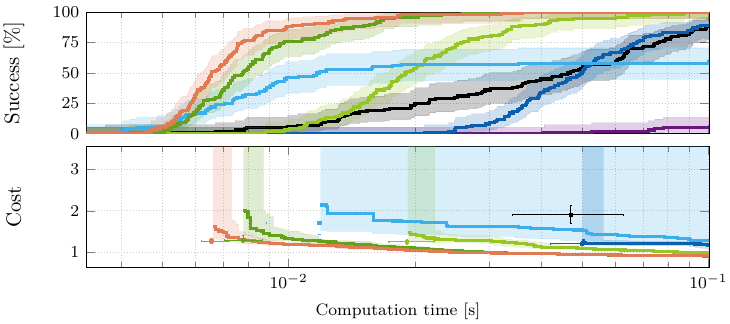}};

    \node[inner sep=0pt] (russell) at (0.0,-4.0){\includegraphics[width=0.8\textwidth]{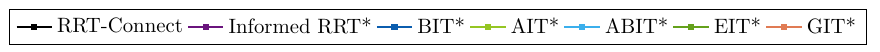}};

    \node at (-4.5,5.8) {\footnotesize (a) Dividing Walls (DW) in $\mathbb{R}^4$ - MaxTime: 0.3s};
    \node at (-4.5,1.3) {\footnotesize (c) Random Rectangles (RR) in $\mathbb{R}^4$ - MaxTime: 0.2s};
    \node at (-4.5,-3.2) {\footnotesize(e) Goal Enclosure (GE) in $\mathbb{R}^4$ - MaxTime: 0.1s};

    \node at (4.5,5.8) {\footnotesize (b) Dividing Walls (DW) in $\mathbb{R}^8$ - MaxTime: 0.6s};
    \node at (4.5,1.3) {\footnotesize(d) Random Rectangles (RR) in $\mathbb{R}^8$ - MaxTime: 1.5s};
    \node at (4.5,-3.2) {\footnotesize(f) Goal Enclosure (GE) in $\mathbb{R}^8$ - MaxTime: 0.5s};
+

    \end{tikzpicture}
    \vspace{-1.0em} 
    \caption{Detailed experimental results from Section~\ref{subsec:experi} are presented above. MaxTime is the planner's maximum allotted planning time. Fig. (a) and (b) depict test benchmark dividing walls outcomes in $\mathbb{R}4$ and $\mathbb{R}^8$, respectively. Panel (c) showcases random rectangle experiments in $\mathbb{R}^4$, while panels (d) demonstrate in $\mathbb{R}^8$. Panel (e) and (f) present goal enclosure experiments in $\mathbb{R}^4$ and $\mathbb{R}^8$. In the cost plots, boxes represent solution cost and time, with lines showing cost progression for optimal planners (unsuccessful runs have infinite cost). Error bars provide nonparametric 99\% confidence intervals for solution cost and time.}
    \label{fig: result}
    \vspace{-1.8em}
\end{figure*}
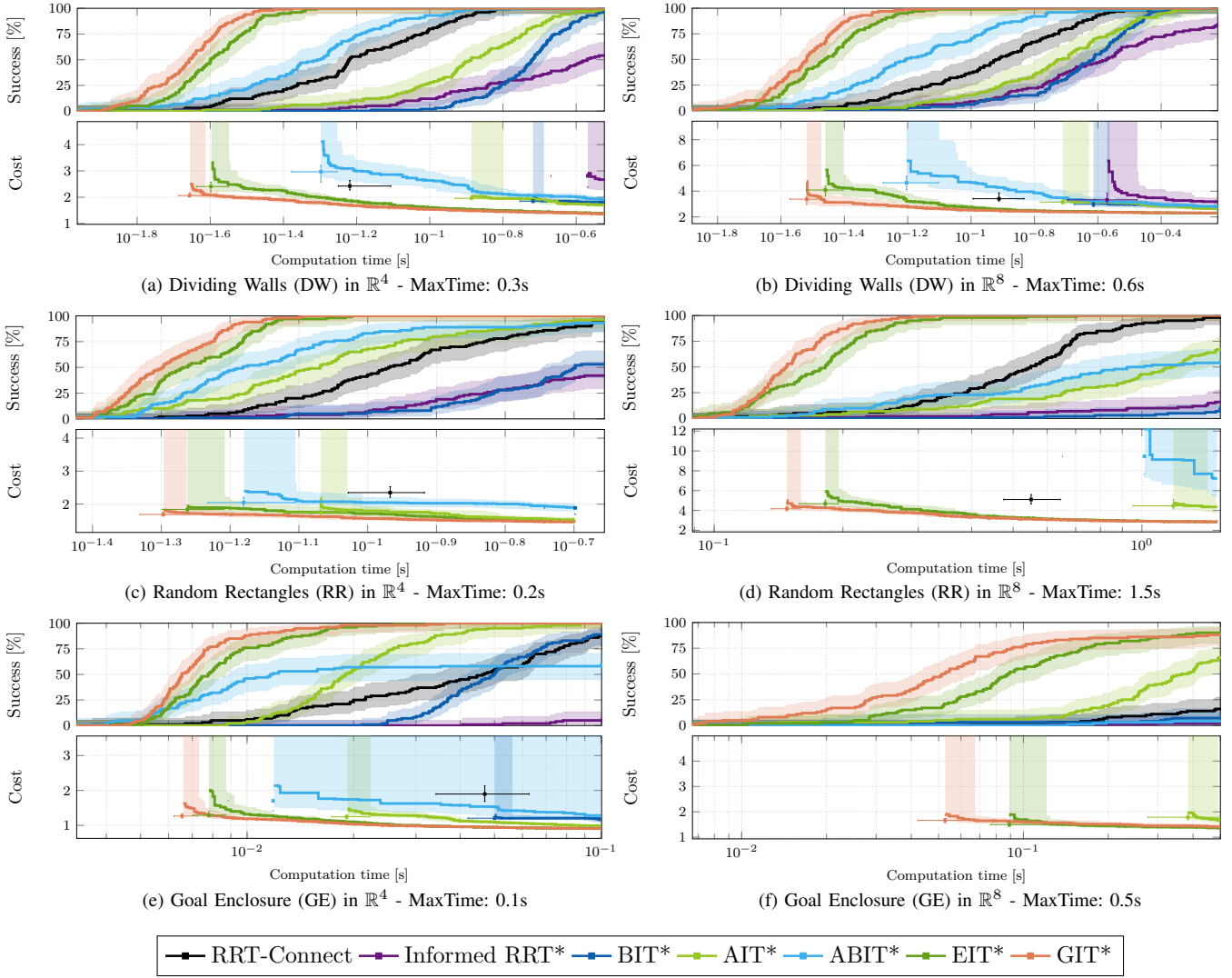
In this paper, we utilize the Planner Developer Tools (PDT)~\cite{gammell2022planner} and MoveIt~\cite{gorner2019moveit} to benchmark motion planner behaviors. GIT* was tested against SOTA algorithms in both simulated random scenarios (Fig.~\ref{fig: testEnv}) and real-world manipulation problems (Fig.~\ref{fig: Realresult}). The comparison involved several versions of RRT-Connect, Informed RRT*, BIT*, AIT*, ABIT*, and EIT* sourced from the Open Motion Planning Library (OMPL)~\cite{sucan2012open}. The evaluations were conducted on a computer with an Intel i7 3.90 GHz processor and 32GB of LPDDR3 3200 MHz memory. {These comparisons were carried out in simulated environments of dimensions $\mathbb{R}^4$ and $\mathbb{R}^8$. The primary objective for the planners was to minimize path length (cost).} The RGG constant $\eta$ was uniformly set to 1.001, and the rewire factor was set to 1.2 for all planners.

In the case of RRT-based algorithms, a goal bias of 5\% was employed, and the maximum edge lengths were determined based on the dimensionality of the space. All batched algorithms utilized a batch size of 100. BIT*, AIT*, ABIT*, and EIT* maintained a linear combination heuristic function of Euclidean distance and effort, respectively. GIT* utilized optimal G-heuristic (Eq.~\ref{equ:key}) to extract the next edge from the reverse queue, which was selected based on the fitness of RGP.

\subsection{Simulation Experimental Tasks}\label{subsec:experi}
The planners were tested across three distinct benchmarks in two domains: $\mathbb{R}^4$ and $\mathbb{R}^8$. In the first scenario, a constrained environment resembling a dividing wall with several narrow gaps was simulated, allowing valid paths in multiple general directions for non-intersecting solutions (Fig.~\ref{fig: testEnv}a). Each planner underwent 100 runs, with computation time for each anytime asymptotically optimal planner shown in the labels, using varying random seeds. The overall success rates and median path lengths for all planners are depicted in Fig.~\ref{fig: result}a and \ref{fig: result}b. It can be seen that GIT* quickly finds the initial solution in both dimensions with minimal time, whereas EIT* requires more time to find the initial solution.

In the second test scenario, random widths were assigned to \textit{axis-aligned hyperrectangles}, generated arbitrarily within the \textit{$\mathcal{C}$-space} (Fig.~\ref{fig: testEnv}b). Random rectangle problems were created for each dimension of the \textit{$\mathcal{C}$-space}, with each planner undergoing 100 runs for every instance. Fig.~\ref{fig: result}c and \ref{fig: result}d illustrate the proposed method has the highest success rates and lowest median path costs within the computation time compared with other planners. This indicates that GIT* can recognize promising regions via environmental information (e.g., APF) where feasible paths likely lie, thereby biasing the sampling process toward these regions. As a result, GIT* outperformed and can quickly find an initial solution.

The last test problem consisted of a hollow, axis-aligned hyperrectangle enclosing the goal state, configured such that even in higher dimensions, the goal can only be reached through the face of the hyperrectangle farthest from the start state (Fig.~\ref{fig: testEnv}c). This problem is challenging for GIT* because there are many invalid edges close to the root of the reverse search tree, often requiring large parts to be repaired (Figs.~\ref{fig: result}e-f). From the figure, the GIT* achieves the best performance in finding the initial solution and converging to the optimal solution compared with the SOTA planner.

As observed in Table \ref{tab:benchmark}, there's a median initial time improvement across varied benchmark scenarios, correlating with dimensionality. For instance, in the $\text{DW}-\mathbb{R}^4$ scenario, GIT* exhibits a lower initial median time (i.e., median value over 100 trials) of 0.0201s compared to 0.0252s for EIT* and 0.1299s for AIT*. This trend is consistent across other scenarios, such as $\text{RR}-\mathbb{R}^4$ and $\text{GE}-\mathbb{R}^4$, where GIT* consistently shows reduced initial median times.

In the $\text{GE}-\mathbb{R}^{8}$ scenario, GIT* demonstrates an initial median time of 0.0512s, compared to 0.0941 for EIT* and 0.3834s for AIT*. This indicates an improvement in initial convergence time of approximately 45.59\% compared to EIT*.

Overall, Table \ref{tab:benchmark} highlights the advantages of GIT* in achieving lower initial median times compared to SOTA, thereby enhancing the efficiency of path planning algorithms.

 \begin{table*}[t]
\caption{Benchmarks evaluation comparison (Fig.~\ref{fig: result})}
\centering
\vspace{-1.0em} 
\resizebox{0.92\textwidth}{!}{
\begin{tabular}{lcccccccccc}
\toprule
 & \multicolumn{3}{c}{${\text{Adaptively Informed Trees}}$}& \multicolumn{3}{c}{${\text{Effort Informed Trees}}$} & \multicolumn{3}{c}{${\textcolor{purple}{\text{Genetic Informed Trees}}}$} &\multirow{2}*{\normalsize$t^\textit{med}_\textit{init}\color{violet}\Uparrow\color{teal}\Uparrow$ (\%)}\\
\cmidrule(lr){2-4} \cmidrule(lr){5-7} \cmidrule(lr){8-10}
    &$t^\textit{med}_\textit{init}$ &$c^\textit{med}_\textit{init}$ &$c^\textit{med}_\textit{final}$ &$t^\textit{med}_\textit{init}$ &$c^\textit{med}_\textit{init}$ &$c^\textit{med}_\textit{final}$ 
    &$t^\textit{med}_\textit{init}$ &$c^\textit{med}_\textit{init}$ &$c^\textit{med}_\textit{final}$ \\
 \toprule

    $\text{DW}-\mathbb{R}^4$   &\textcolor{violet}{0.1299}   &1.9571   &1.7151 &\textcolor{teal}{0.0252} &2.4051 &1.3693 &\textcolor{purple}{0.0201} &2.0619 &1.3634 &\textcolor{violet}{{84.53}} / \textcolor{teal}{20.23}  \\
    $\text{DW}-\mathbb{R}^8$   &\textcolor{violet}{0.1947}   &3.1492   &2.6388 &\textcolor{teal}{0.0357} &{4.0910} &2.2892  &\textcolor{purple}{0.0279} &{3.3791} &2.3109 &\textcolor{violet}{{85.67}} / \textcolor{teal}{21.84}  \\
 \midrule
    $\text{RR}-\mathbb{R}^{4}$   
    &\textcolor{violet}{0.0853} &{1.7570} &{1.5282} &\textcolor{teal}{0.0587} &{1.8392} &{1.4715}  &{\textcolor{purple}{0.0472}} &{1.6874} &{1.4595} &\textcolor{violet}{44.67} / \textcolor{teal}{19.59}\\   
    $\text{RR}-\mathbb{R}^8$    &\textcolor{violet}{1.1843} &4.4697 &4.3599 &\textcolor{teal}{0.1889} &4.6789 &2.8588 &{\textcolor{purple}{0.1429}} &4.1716 &2.8450  &\textcolor{violet}{\textbf{87.93}} / \textcolor{teal}{24.35}\\
 \midrule
    $\text{GE}-\mathbb{R}^4$  &\textcolor{violet}{0.0191}  &1.2457  &0.9900 &\textcolor{teal}{0.0082}  &1.2909  &0.9126
    &{\textcolor{purple}{0.0064}} &1.2678 &{0.9083} &\textcolor{violet}{66.49} / \textcolor{teal}{21.95}  \\
    $\text{GE}-\mathbb{R}^{8}$   &\textcolor{violet}{0.3834}   &1.7854   &1.6605 &\textcolor{teal}{0.0941}   &1.4970   &1.4086 &{\textcolor{purple}{0.0512}} &{1.6634} &{1.3636}  &\textcolor{violet}{{86.64}} / \textcolor{teal}{\textbf{45.59}}  \\
\bottomrule
\end{tabular}} \label{tab:benchmark}
\vspace{-1.7em} 
\end{table*}
\subsection{Real-world Path Planning Tasks}\label{subsec:realExpri}
\begin{figure*}[t!]
    \centering
    \begin{tikzpicture}
    \node[inner sep=0pt] (russell) at (0,0)
    {\includegraphics[width=0.94\textwidth]{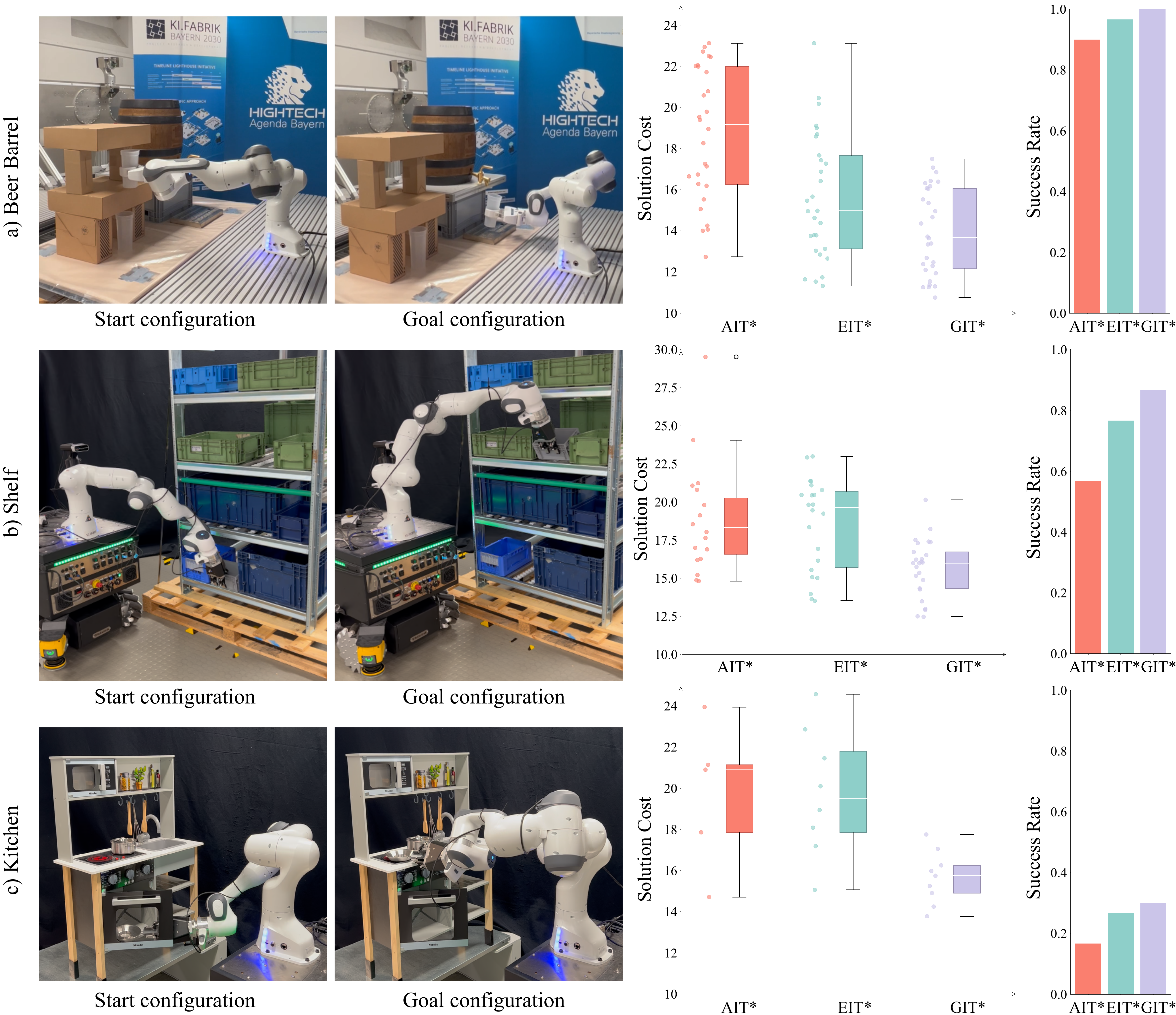}};

    \end{tikzpicture}
    \vspace{-0.3em} 
    \caption{Detailed experimental results from Section~\ref{subsec:realExpri} are summarized above. Fig.~\ref{fig: Realresult}a illustrates the \textit{beer barrel} ENV, highlighting the start and goal configurations along with the solution cost and success rate. Fig.~\ref{fig: Realresult}b depicts the \textit{industry shelf} ENV, showing the initial and final positions for extracting and placing an industry-standard container. Fig.~\ref{fig: Realresult}c presents the \textit{kitchen} ENV, focusing on the DARKO robot's performance. In the cost box plots, boxes indicate the solution cost per planner, while lines represent the mean cost progression for an optimal planner (unsuccessful runs are assigned an infinite cost).}
    \label{fig: Realresult}
    \vspace{-1.7em}
\end{figure*}
To evaluate the algorithm's performance in real-world scenarios, three numerical experiments are conducted on a single-arm manipulator and mobile manipulator (DARKO) to demonstrate the efficiency and extensibility of GIT* compared with three SOTA path planning algorithms: Batch Informed Trees (BIT*)~\cite{gammell2020batch}, Adaptively Informed Trees (AIT*)~\cite{strub2022adaptively}, and Effort Informed Trees (EIT*)~\cite{strub2022adaptively}. 

We compare GIT* with AIT* and EIT* in single-arm manipulator environments to evaluate their performance in converging to the optimal solution cost and success rate over 30 runs. The first environment (\textit{Beer Barrel}) consists of simple cup holder obstacles. The second and third environments (\textit{Shelf} and \textit{Kitchen}) are confined to the DARKO robot and cluttered with narrow spaces. A collision-free path connecting the start state to the goal region is required. GIT* demonstrated its effective G-heuristic during multiple experimental tasks (Fig.~\ref{fig: Realresult}). The detailed behavior of real-world experiments can be viewed in the accompanying video.
\subsubsection{\textbf{Beer Barrel Cup Placement Task}}
Fig.~\ref{fig: Realresult}a showcases the start and goal configuration of the cup placement task. In this task, we utilize a single robotic manipulator to grab a beer cup and place it under the beer tap of the beer barrel keg while avoiding obstacles. The following graph illustrates the performance of AIT*, EIT*, and GIT* in terms of solution cost and success rate.
All planners were given 1.0 seconds to address the beer barrel cup placement problem. Over the course of 30 trials, GIT* achieved a 100\% success rate with a median solution cost of 13.8972. EIT* had a success rate of 96.67\% with a median solution cost of 15.1332. AIT* was 93.33\% successful, with a median solution cost of 19.2183.

\subsubsection{\textbf{Industry Shelf Container Rearrangement Task}}
The initial and final configurations for the shelf task are depicted in Fig.~\ref{fig: Realresult}b. This task involves extracting an industry-standard container from a position between two other boxes on the lower bottom layer and repositioning it on the third layer of the shelf, again between two containers. Due to component standardization, the challenge lies in the precise insertion of industry containers into narrow spaces. The task aims to place the industry-standard container between two larger containers on the shelf, with a tolerance scope of $\leq$5mm, making the planning of a collision-free feasible path particularly difficult.
Each planner was allocated 5.0 seconds to solve this confined, limited space pull-out and insertion problem. Across 30 trials, GIT* achieved an 86.67\% success rate with a median solution cost of 15.9745. EIT* had a 76.67\% success rate with a median solution cost of 19.1045. AIT* managed a 56.67\% success rate with a median solution cost of 18.2672.

\subsubsection{\textbf{Kitchen Model Pan Cooking Task}}
For the third task, we utilized the DARKO robot positioned in front of a kitchen model. The start and goal configurations are illustrated in Fig.~\ref{fig: Realresult}c. This task is particularly challenging as the manipulator must navigate the geometric shape of the pan within a cluttered oven while also avoiding collisions between the base robot and the kitchen shelves. The complexity is further heightened by the need for precise movements in a confined space.
Each planner was allotted 10.0 seconds to solve this kitchen pan reallocation problem. Over the course of 30 trials, GIT* achieved a 30\% success rate with a median solution cost of 15.8860. EIT* had a success rate of 26.67\% with a median solution cost of 19.2746. AIT* managed a 16.67\% success rate with a median solution cost of 20.9824.

In short, compared with the AIT* and the EIT*, the GIT* achieves the best performance on finding the initial solution and converging to the optimal solution.
\subsection{Discussion}
\subsubsection{Comparison With SOTA Planner} {To showcase the advantages of GIT*, we compared its performance with AIT* and EIT* using success rate and solution cost metrics in three real-world tasks (Fig.~\ref{fig: Realresult}): placing cups on beer barrel faucets, rearranging industrial containers on shelves, and cooking pans in a kitchen model, and six simulation tasks  (Fig.~\ref{fig: testEnv}) across multi-dimensions with randomly generated seeds.}

From the experiment results, we observe that EIT* performs much better than AIT* in both simulation environments (Table~\ref{tab:benchmark}) and real-world scenarios. However, GIT* outperforms SOTA planners due to its use of problem-specific environmental information via RGP and the integration of the G-heuristic. As shown in Fig.~\ref{fig: result}(a, c, and e), In low-dimensional problem domains, the initial solution finding time and cost show minimal improvement. In high-dimensional domains, the linear combination heuristic struggles to guide the search efficiently, as shown in Fig.~\ref{fig: result}(b, d, and f). Furthermore, In the first real-world environment, GIT* outperformed EIT* by 3.33\% in success rate and reduced the solution cost by approximately 8.17\%. Compared to AIT*, GIT* improved the success rate by 6.67\% and reduced the solution cost by approximately 27.68\%, as shown in Fig.~\ref{fig: Realresult}(a).
In the second real-world experiment, the benchmark results show that the G-heuristic can enhance solving cluttered tasks, achieving the highest success rate and the lowest solution cost among the evaluated planners. GIT* outperformed EIT* by 13.04\% in success rate and reduced the solution cost by approximately 16.38\%. Compared to AIT*, GIT* improved the success rate by 52.94\% and reduced the solution cost by approximately 12.56\%, as shown in Fig.~\ref{fig: Realresult}(b).
In the third real-world experiment, GIT* outperformed EIT* by 12.5\% in success rate and reduced solution cost by about 17.58\%. Compared to AIT*, GIT* improved success rate by 80\% and reduced solution cost by approximately 24.32\%, as shown in Fig.~\ref{fig: Realresult}(c). These results highlight the effectiveness of the G-heuristic in narrow environments to prevent obstacle avoidance in the kitchen model.

From the discussion, one may conclude that using RGP to train an optimal G-heuristic across all benchmarks can improve the initial convergence rate and initial path length. Furthermore, GIT* can utilize environmental information to search via more promising regions (i.e., APF and dynamic importance), which accelerates the path-planning initial finding process. GIT* achieved the highest success rate and lowest solution cost among tested SOTA planners, emphasizing its potential for real-world applications.

\subsubsection{Limitations and Future Work} While GIT* demonstrates superior performance, it has limitations. The current implementation is tailored for specific tasks with predefined start and goal configurations, limiting its adaptability to variable environments and tasks. Future work could enhance GIT*'s generalization capabilities by integrating neural network-driven approaches to learn from diverse human demonstrations, improving its extension ability across different environments. This aligns with advancements in neural network-based path planning and promises to enhance GIT*'s robustness and versatility. Furthermore, future designs will consider human acceptability and comfort when planning trajectories.

\section{Conclusion}
In this paper, we introduced the Genetic Informed Trees (GIT*) algorithm, a novel path planning approach that leverages Reinforced Genetic Programming (RGP) to refine heuristic functions for enhanced guidance. By incorporating additional environmental data, such as repulsive forces from obstacles and the dynamic importance of vertices, GIT* improves search efficiency and solution quality. The integration of RGP allows GIT* to mutate genotype-generative heuristic functions (G-heuristic), adapting to various problem domains. Our comparative analyses demonstrate that GIT* consistently outperforms existing single-query, sampling-based planners across different scenarios, including simulation benchmarks and real-world robot manipulation tasks. Optimal G-heuristic exhibits notable improvements over SOTA methods in terms of both success rate and solution cost, showcasing its robustness and adaptability, particularly in handling complex, cluttered environments with high precision and efficiency.

In conclusion, GIT* enhances rapid initial pathfinding and reduces solution costs. GIT* shows promising potential for future research and applications in motion planning.

\bibliographystyle{IEEEtran}
\bibliography{bibliography}

\end{document}